%% file: casual-narrative.tex
\documentclass[preprint,journal]{vgtc}                

\ifpdf
  \pdfoutput=1\relax                   
  \usepackage{graphicx}                
  \DeclareGraphicsExtensions{.pdf,.png,.jpg,.jpeg} 
\else
  \usepackage{graphicx}                
  \DeclareGraphicsExtensions{.eps}     
\fi%

\graphicspath{{figures/}{pictures/}{images/}{./}} 

\usepackage{microtype}                 
\PassOptionsToPackage{warn}{textcomp}  
\usepackage{textcomp}                  
\usepackage{mathptmx}                  
\usepackage{times}                     
\usepackage{cite}                      
\usepackage{tabu}                      
\usepackage{booktabs}                  
\usepackage[table,xcdraw]{xcolor}
\usepackage{multirow}
\usepackage{Carrickc}
\usepackage{lettrine}

\usepackage[ruled,vlined,linesnumbered,algo2e]{algorithm2e}
\usepackage{enumitem}
\usepackage{subfig}
\usepackage{svg}
\usepackage{hyperref}

\usepackage{xcolor}

\usepackage{amssymb}

\onlineid{1136}

\vgtccategory{Research}

\vgtcpapertype{Algorithm/Technique}

\newcommand{\thetitle}{Once Upon A Time In Visualization: Understanding the Use of Textual Narratives for Causality} 
\title{\thetitle}

\author{Arjun Choudhry, Mandar Sharma, Pramod Chundury, Thomas Kapler, Derek W.\ S.\ Gray, \and Naren Ramakrishnan, and Niklas Elmqvist, \textit{Senior Member, IEEE}}

\authorfooter{
\item Arjun Choudhry, Mandar Sharma, and Naren Ramakrishnan are with
Virginia Tech in Arlington, VA, USA.
Email: \{aj07lfc, mandarsharma\}@vt.edu, naren@cs.vt.edu 
\item Pramod Chundury and Niklas Elmqvist are with the University of Maryland, College Park, MD, USA.
Email: \{pchundur,elm\}@umd.edu
\item Thomas Kapler and Derek W. S. Gray are with Uncharted Software in Toronto, ON, Canada.
Email: \{tkapler,dgray\}@uncharted.software
}

\shortauthortitle{Once Upon A Time In Visualization}
\abstract{Causality visualization can help people understand temporal chains of events, such as messages sent in a distributed system, cause and effect in a historical conflict, or the interplay between political actors over time.
  However, as the scale and complexity of these event sequences grows, even these visualizations can become overwhelming to use.
  In this paper, we propose the use of textual narratives as a data-driven storytelling method to augment causality visualization. 
  We first propose a design space for how textual narratives can be used to describe causal data.
  We then present results from a crowdsourced user study where participants were asked to recover causality information from two causality visualizations---causal graphs and Hasse diagrams---with and without an associated textual narrative.
  Finally, we describe \textsc{CauseWorks}, a causality visualization system for understanding how specific interventions influence a causal model.
  The system incorporates an automatic textual narrative mechanism based on our design space.
  We validate \textsc{CauseWorks} through interviews with experts who used the system for understanding complex events.
}

\keywords{Causality visualization, natural language generation, data-driven storytelling, temporal data, quantitative studies.}

\begin{document}

\newcommand{\change}[1]{}
\newcommand{\aj}[1]{{\color{black}\textrm{#1}}}
\newcommand{\ms}[1]{{\color{black}\textrm{#1}}}

\firstsection{Introduction}

\maketitle

\input{01-intro}

\input{02-background}

\input{03-design-space}

\input{04-formative}

\input{05-application}

\input{06-discussion}

\input{07-conclusion}

\acknowledgments{This work was partially supported by the Defense Advanced Research Projects Agency (DARPA) under Contract Number FA8650-17-C-7720.
The views, opinions and/or findings expressed are those of the authors and should not be interpreted as representing the official views or policies of the Department of Defense or the U.S.\ Government.
The authors wish to thank all DARPA Causal Exploration collaborators for their support and encouragement.}

\bibliographystyle{abbrv-doi}
\bibliography{casual-narrative}

\newpage
\input{appendix}

\end{document}

%% file: 01-intro.tex
\label{sec:intro}

\lettrine[lines=3]{S}{tories} are a central part of what it means to be human~\cite{Leitch1986, Schank1995}.
They teach, guide, and caution; they store, recall, and archive; they praise, spread joy, and inspire.
In particular, stories are especially useful for encapsulating \textit{causality}---the cause and effect of events in a plot---in a regular, understandable, and memorable format.
This format is also surprisingly scalable. 
Examples abound of textual narratives representing complex chains of cause and effect ranging from the winding plots of G. R. R. Martin's \textit{A Song of Ice and Fire} and Neal Stephenson's \textit{The Baroque Cycle}, through shelf yards of history books laying out the intricacies of the Napoleonic Wars or the American Revolution in all their gritty detail, and all the way to quarterly reports telling the story of a company's accomplishments over the last three months.
However, despite all of this utility, little work exists on the use of textual narratives to represent causality in modern visualization tools.
On the contrary, visualization and visual analytics researchers tend to view textual narratives with suspicion, often instead opting to apply text analytics and visualization methods to minimize their use.

In this paper, we attempt to remedy this gap in the literature by investigating how textual narratives can be used to represent causality.
Our classification of narratives is primarily based on their utility as a complement to causality visualization techniques, such as dynamic graphs~\cite{Beck2014} and Hasse diagrams~\cite{Elmqvist2003a, Elmqvist2003b}.
Textual representations are generally much less compact than geometric ones (i.e., visualizations), and must thus be designed with specific questions in mind.
We first \change{change 1}\aj{propose and discuss a} design space of causality representations, focusing in particular on textual narratives. \change{change 2}\aj{We then report on a crowdsourced user study where we operationalized parts of our design space and asked participants to} recover causality information from dynamic graphs versus Hasse diagrams, with and without an associated textual narrative.
Our findings indicate that narratives can fill an important complementary role for key questions on causality, and thus serve as a ``story-like'' format to summarize a specific causal event chain.

To capitalize on these findings \change{change 3}\aj{and demonstrate the use of our design space}, we also present a textual narratives implementation in \textsc{CauseWorks}, a causality visualization system on the Causal Exploration of Complex Operational Environments program~\cite{causeex} for understanding the impact of specific interventions in a causal model.
These narratives are based on best practices from our design space \change{change 4}\aj{as well as the user study}, and serve as a quick-reference textual summary of the selected interventions and objectives shown in a dynamic graph.
We studied the utility of these narratives by interviewing several users with experience of causality, who used the system to understand climate change data. 
Our findings confirm many of the results from the crowdsourced study.

The contributions of our paper are the following: (1) a design space for complementary textual narratives in representing causality; (2) results from a crowdsourced study evaluating different causality visualizations with and without companion narratives; (3) an implementation of textual narratives in an existing causality analytics and visualization system (\aj{\textsc{CauseWorks}}); and (4) qualitative results from 5 experts using these narratives to understand climate change data.

%% file: 02-background.tex
\section{Background}
\label{sec:background}

\aj{Here we discuss the existing literature on causality, causality visualization, and data-driven storytelling.}

\subsection{Causality Visualization}

Causal networks or directed acyclic graphs are commonly used to map relationships between variables~\cite{Pearl2000CausalityMR}.
Much of the work in causal visualization aims to encode aspects of causality such as temporal developments of cause and effect~\cite{Elwert201313GC} using interactivity~\cite{7192729, ware2013} and animations~\cite{kadaba2007visualizing} to improve the accuracy of causal inference.

Researchers have uncovered characteristics and shortcomings of specific visual representations of causality.
For example, Bae et al.~\cite{bae2017understanding} find that multiple to/from connections from a particular node may influence how an analyst perceives indirect effects.
Similarly, Hasse diagrams are used widely, but require the user to backtrace every effect, and can also introduce an overwhelming number of crossings in a large-scale causal system~\cite{elmqvist2004animated}.
Wang et al.\ attempted to improve causal inference by overlaying salient statistical parameters such as p-values and regression co-efficients on 2D-graphs so that analysts can draw more reliable conclusions about causal relationships~\cite{7192729}.
However, interpreting these parameters requires understanding statistical inference.

Research on perceptions of causality also show that inference is context-dependent, and a non-expert with regards to statistics or domain could see an illusion of causality in data~\cite{xiong2019illusion}.
In our work, we propose to mitigate misinterpretation by both experts and non-experts alike through the use of textual narratives to augment causal visualizations.

\subsection{Narratives in Visualization}
 \aj{Historically spanning thousands of years~\cite{Schank1995, Vansina1985}, \textit{storytelling} conveys a series of events, usually involving characters and locations---\textit{stories}---using speech, sound, and visuals~\cite{Gottschall2012}. Generally, stories are communicated using visual media, such as illustrations, pictures, animations, video, and--now--visualization~\cite{Eisner2008, Sless1981}. Visualization, inherently, is inclined for communication by virtue of its graphical form, resulting in the notion of \textit{communication-minded visualization}~\cite{Viegas2006}. Combining the idea of \textit{communication-minded visualization} with \textit{storytelling} yields the notion of \textit{data-driven storytelling}: narrative techniques for data~\cite{Segel2010}.}
 
 \aj{We believe that data-driven storytelling naturally follows the idea of visualization for explanation (the latter). The production, presentation, and dissemination of analysis results is an important challenge in visualization and visual analytics~\cite{Thomas2005}. Gershon and Page first proposed using storytelling for visualization~\cite{Gershon2001}, and their work has since been followed up by workshops~\cite{Diakopoulos2011, DiMicco2010}, surveys~\cite{Hullman2011, Segel2010}, and even commercial tools~\cite{Kosara2013}. Vi{\'e}gas and Wattenberg note the inclination of visualization for communication by virtue of its graphical form, and encourage focusing on so-called \textit{communication-minded visualization}~\cite{Viegas2006} for social analysis. In recent years, the use of textual data to aid visualization and vice versa have been explored~\cite{8440852, 7864462}. Furthermore, verbalization~\cite{Sevastjanova2018Going-45045, 8933695, 10.1111:cgf.14034} has also been used for understanding machine learning models.}

\vspace{-0.15cm}
\subsection{Causality and Causal Networks}

The statistical and ML sciences have developed many formalisms to reason with both the structure and dynamics of causal networks~\cite{Bunge1979CausalityAM}. 
To encapsulate causal structure, while there are many network formulations, one of the more popular ones is the Bayesian network formalism popularized by Pearl~\cite{Pearl2000CausalityMR}.
A Bayesian network is a directed acyclic graph (DAG) and can be thought of as a way to represent a factorization of the underlying joint distribution of random variables.
However, interpreting such DAGs \aj{is} difficult for humans and interpretation rules such as d-separation\cite{geiger1990d} and the `Bayes Ball' algorithm \cite{shachter2013bayes} have been proposed.
These rules essentially are ways to read or infer conditional independence relationships from the networks.

To overcome such interpretation difficulties, other representational formalisms have been proposed, e.g., dependency networks~\cite{dep-networks}\aj{,} which allow cycles, and Markov networks~\cite{koller-friedman} (also called Markov random fields, or MRFs), which are undirected.
In terms of dynamics, a causal representation must allow us to probe the effect of interventions and to posit and explore counterfactuals.
Interventions are modeled using a calculus (e.g., Pearl's do-calculus) that mutates the given network to propagate and understand the downstream consequences of the intervention.
Counterfactuals allow us to ask more expressive questions and explore the progression of different variables in alternative worlds or situations.
We assume in this paper that the underlying causal representation is fixed and a suitable interpretation of dynamics is available to probe the effect of interventions, and focus on the role of visualization in communicating cause-effect relationships.

%% file: 03-design-space.tex
\vspace{-0.2cm}
\section{Design Space: Textual Narratives for Causality}
\label{sec:design-space}

Visualizations are themselves considered as ways to tell stories with data~\cite{Segel2010} and, in this paper, we view textual narratives as an augmented form of storytelling that aims to increase insight\cite{chang2009defining, north2006toward}, comprehension, and decision making.
We focus on textual narratives as a way to express causal information in event sequences, specifically as a complement to causality visualizations\cite{hullman2013deeper, moezzi2017using}, such as causal graphs or Hasse diagrams.
For this reason, we tend to think of these textual narratives as a form of \textit{data-driven storytelling}~\cite{Henry2018}---the use of traditional narrative methods to convey data---that relies on a textual, rather than a visual, medium.
We believe that a textual narrative can also replace the visualization, at least to provide a high-level summary\change{change 7}\aj{~\cite{Nenkova2012ASO, Gambhir2016RecentAT}}.

\vspace{-0.15cm}
\subsection{Definitions}
\label{ss:def}

The causal relation $\rightarrow$ is a relation that connects two elements (events) $x$ and $y$ as $x \rightarrow  y$ iff $x$ is the cause of $y$.
Sets of events are called processes $P_1, \ldots, P_N$. 
Internal events are sequential and causally related.
External events interconnect processes through messages.
We denote the events for a process $P_i$ as $E_i = \{ e^i_1, e^i_2, e^i_3, \ldots \}$.
The causal relation is typically irreflexive, asymmetric, and transitive.

While some causal tasks are concerned with the entire causal model---i.e., the set of all processes and their associated events---many real-world tasks 
use a more directed formulation.
Of most interest is the ability to impose specific \textit{interventions} or perturbations on a particular process and understanding the resulting impact on \textit{objective} process(es).
A causal model may involve a set of interventions $I$ and objectives $O$, each corresponding to specific causes and effects.

\vspace{-0.1cm}
\subsection{Narrative Rendering Pipeline}
\label{sec:pipeline}
We view the representation of causal data using textual narratives as an interactive \textit{rendering pipeline}, akin to a classic graphics rendering or visualization pipeline.
In this model, we think of the \textit{sentence clause} as the building block. 
Natural language generation (NLG) systems~\cite{Dale2000, nlg2} tend to consist of several stages:

\begin{enumerate}
    \setlength{\itemsep}{0pt}
    \item\textbf{Content selection:} Determining the causality data to display;
    \item\textbf{Document structuring and aggregation:} Prioritizing the order of data and merging sentences on similar causal data (same source or destination processes) to improve readability;
    \item\textbf{Realization:} Generating the actual text for each piece of causality information to render in the summary; and
    \item\textbf{Interaction:} Providing a feedback loop to allow interaction with the textual narratives, such as to drill down, link to related narratives, or brush to highlight items in associated views.
\end{enumerate}
 
Using natural language to represent data as text is quite different from using visualization, which uses geometric shapes. 
Unlike visualization, natural language is typically precise.
This leads to early fixation, as well as serial representations, which limits parallel processing.
In practice, this means that natural language is better suited to presenting specific pieces of information rather than the holistic and parallel overviews that characterize data visualization.

\subsection{Step 1: Extracting Causality Information}\label{sec:stepCausalityExtraction}

Our proposed text generation pipeline starts with identifying the specific causality information that users desire.
This generally depends on the application, which we model using a degree-of-interest (DOI) function in the next language generation step. 
Thus, our treatment here includes all potentially useful causality information, given our data model.
We organize this information into the following categories:

\begin{itemize}
    \setlength{\itemsep}{0pt}

    \item\textbf{Cause and effect:} A central question when reasoning about causality tends to be the factors that caused a specific effect, which we capture as interventions and objectives. \textit{Example:} a white cue ball striking the eight ball, sending it bouncing off the nearest wall of a pool table.

    \item\textbf{Correlation:} While correlation is not causation, many forms of causation have their roots in correlation.
    Depending on the causality model, the exact cause and effect may not be known; in such cases, correlations between nodes---i.e., a change in one node followed by a change in another node---can be used as a weaker form.
    \textit{Example:} a medication administered to a patient followed by their blood pressure dropping.

    \item\textbf{Life cycle:} Processes may come and go, often as a result of them receiving an intervention or an internal event. 
    Such life cycle information is commonly of interest in causal reasoning.
    \textit{Example:} traffic in a computer network being directed around a faulty router that is no longer responding.

    \item\textbf{Connectivity:} Causality modeled as above is essentially a graph, which means that understanding a causal model requires understanding the topology and dynamic connectivity of the events passed in the system.
    \textit{Example:} tracing infections of an airborne virus in a population based on their social contacts.

\end{itemize}

\begin{figure}[ht]
\vspace{-0.25cm}
    \centering
    \includegraphics[width=\linewidth]{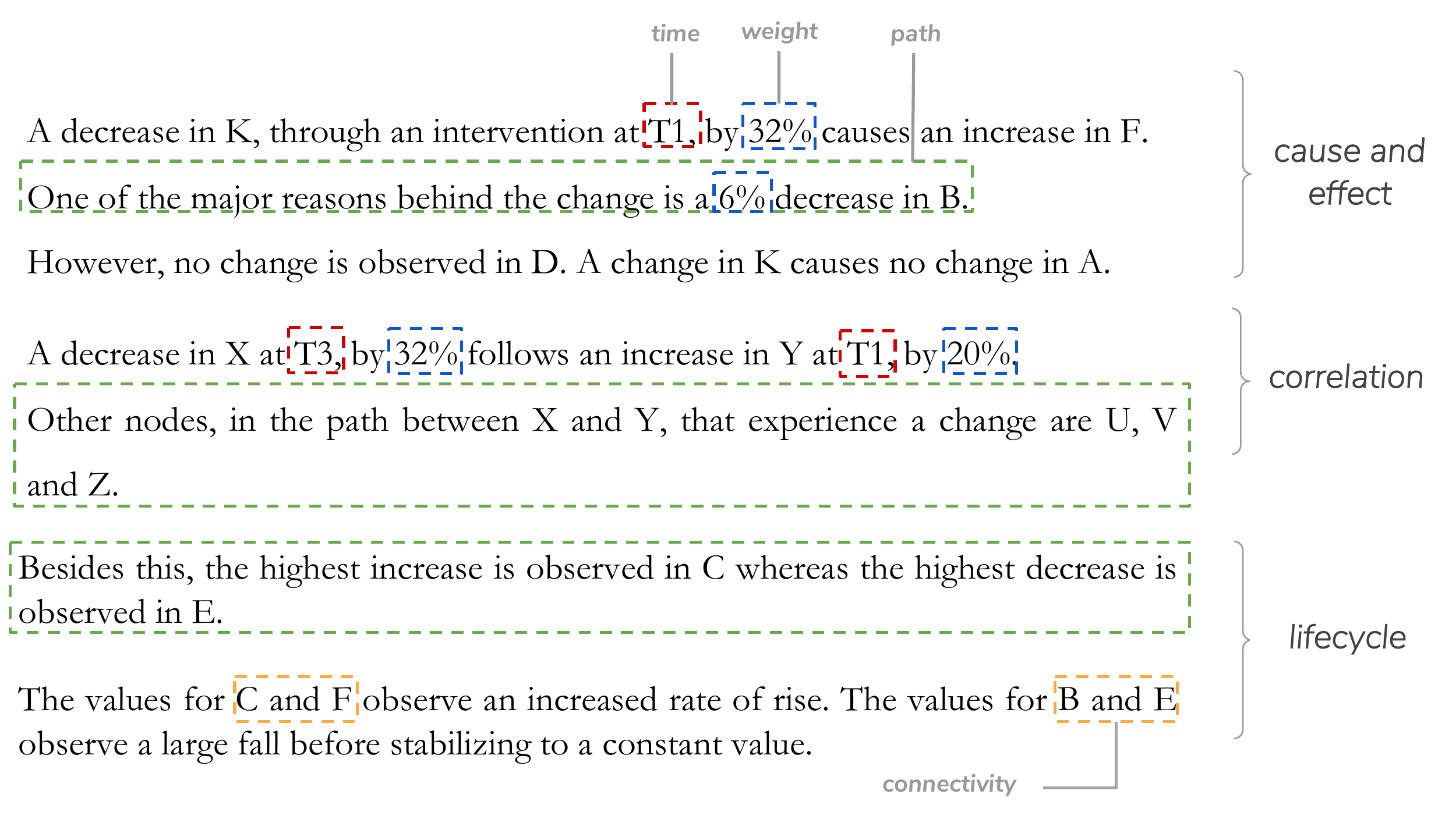}
    \caption{Annotated example of a narrative conveying causal information about interventions and objectives.}
    \label{fig:narrative-example}
\end{figure}

For all of the above causality information categories, we can also identify specific common metadata for them all:
\textbf{path:} the processes on the path between source and destination; \textbf{weights:} the values or weights associated with each process; and \textbf{time:} the time stamps associated with each of the events. \change{change 8}\aj{The types of causal information described above are included in the generated narratives of the crowdsourced study and in \textsc{CauseWorks}  (figure~\ref{fig:narrative-example} shows an example).}

\subsection{Step 2: Calculating Order}\label{sec:stepCalculatingOrder}

Here we determine the structure and order of information that we will use for the textual narrative. 
The primary challenge is that even a moderately complex causal system will have a significant number of candidate causal information to convey.

To address this challenge, we use a degree-of-interest (DOI) function $f_{DOI}(e) \subseteq \mathbb{R}$ based on user interest and task to prioritize each event $e$ involved in the sequence of causality data extracted from the prior step.
As a first level of prioritization, we propose limiting reports to the sets of interventions $I$ and objectives $O$, as described in Section~\ref{ss:def}.
We group items on a per-process basis, and then further prioritize events based on their occurrences, magnitude of change, and influence.
To represent this information, we use a directed acyclic graph (DAG) as a scene graph to store the abstract data to render, where each causal process becomes a top-level container for associated causality data.

Generally speaking, generating a complete sentence for each clause---recall that a clause corresponds to an individual item of data---is the most clear and unambiguous approach.
However, this often leads to significant repetition, which is often seen as clumsy and unnatural to a reader, as well as unnecessary verbosity, which is wasteful given that summaries are often limited in length.
For this reason, we use \textit{aggregation} to merge similar clauses that share the same source or destination process into a single sentence. \change{change 9}\aj{We have aggregated events in our generated narratives (see Figures~\ref{fig:task} and \ref{comparison}).}

\subsection{Step 3: Rendering Textual Narratives}\label{sec:renderingNarratives}

We think of realizing the ordered causality data to be expressed as \textit{rendering} the narrative, akin to how a computer graphics system may render a sorted list of triangles to generate a 3D scene.
Since our focus is on generating summaries, the notion of a \textit{character budget} is central to our approach: this is the maximum number of characters that we want to use to realize the textual narrative.
This budget is not prescriptive, only restrictive; in other words, if space is not an issue, the budget can be set to infinity, resulting in exhaustive textual summaries.

The actual rendering process proceeds by iterating through the sorted data graph, where items are grouped based on top-level processes, as described above.
By knowing the number of characters for each branch of the data graph, the renderer can determine how deeply to traverse while maintaining the character budget.
Furthermore, we can also identify the available visual channels for conveying data using text:

\begin{itemize}
    \setlength{\itemsep}{0pt}
    \item\textbf{Textual content:} The primary visual channel is obviously the written content that the text spells out.
    
    \item\textbf{Font size:} Most summaries will use a uniform font size, as changing the size of individual words or sentences throughout a text can be disruptive to reading as well as when calculating its space needs.
    However, it can be an effective way to show emphasis, particularly for titles and section headings.
    
    \item\textbf{Typographic emphasis:} A more common and typographically accepted practice is to use emphasis such as \textbf{boldface}, \textit{italics}, or \underline{underlining} to communicate data in the narrative, such as to mark processes, effects, or magnitudes. 
    Additional such emphasis markers include \textsc{Small Caps}, \uppercase{All Caps}, or the use of punctuation (!) or ``quotation marks'' in the text.
        
    \item\textbf{Color:} As for visualization, color can be an effective visual channel.
    We differentiate between the use of \textcolor{green!60}{font color} and \colorbox{blue!30}{background color}, which can be used to convey different data (although, as always, care must be taken to avoid interference).
        
    \item\textbf{Hierarchical lists:} While not part of classic running prose, which tends to just use sentences and paragraphs as its typographic structures, we also consider lists---both enumerated and itemized ones---a useful visual channel.
    In particular, nesting lists can allow for showing hierarchical part-of relationships. 
    
    \item\textbf{Word-scale graphics:} Despite our focus on written text, we cannot resist drawing on visualization, in the form of \textit{word-scale graphics}~\cite{Goffin2017}: small data-driven graphics that can be embedded into running text.
    Examples include mathematical symbols such as $\uparrow, \bigtriangleup$, and $\bowtie$, icons such as \includegraphics[height=0.8\baselineskip]{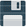}, \includegraphics[height=0.8\baselineskip]{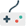}, and \includegraphics[height=0.8\baselineskip]{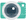}, as well as micro visualizations such as \includegraphics[height=0.8\baselineskip]{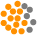}, \includegraphics[height=0.8\baselineskip]{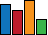}, and \includegraphics[height=0.8\baselineskip]{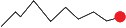} (sparklines).
\end{itemize}

\subsection{Step 4: Interacting with Narratives}\label{sec:interactingNarratives}

Finally, since our intended output format almost always is on a computer screen---and not paper---we should also consider how to interact with these textual narratives. 
We propose the following possibilities:

\begin{itemize}
    \setlength{\itemsep}{0pt}
    \item\textbf{Brushing:} Hovering over a process or an event in a narrative highlights all of its occurrences in related views (or narratives).
    
    \item\textbf{Hyperlinking:} Entities in the narrative are hyperlinks where clicking on one will navigate to it in related views (or narratives).
    
    \item\textbf{Drill-down/roll-up:} Dynamically changing the user's degree of interest will allow for drilling down, e.g., by unfolding the items in a list or expanding suppressed elements in an enumeration.
    
    \item\textbf{Search:} Directly querying specific elements in a narrative by typing partial or complete search terms allows for quick access~\cite{Feng2018}.

\end{itemize}

\vspace{-0.3cm}

\subsection{Discussion: What Makes an Effective Narrative?}

Textual narratives are slowly making their way into visualization systems, either as a way to generate data insights to accompany a visualization~\cite{endert-vast} or to structure a visualization for better communication~\cite{fromendert}.
Research into what makes an effective narrative is still in its infancy and is necessarily tied to the underlying analytical task and domain.
For the causal networks domain considered here we identify four facets:
\vspace{-0.08cm}
\begin{itemize}
    \setlength{\itemsep}{0pt}
    \item \textbf{Language diversity:}
    More language diversity avoids monotony but might detract from conveying key messages and conclusions.
    Less language diversity supports comparison of generated narratives but might lead to `glossing over' by analysts. 
    
    \item \textbf{Level of detail:} Should the narrative capture an executive summary or provide in-depth access to the underlying data?\change{change 10}\aj{ We briefly discuss the preferred level of detail in our expert review.}
    
    \item \textbf{Verbalizing numbers:} Verbalizing quantitative/probabilistic data (e.g., using Kent's words of estimative probability~\cite{kent} or the NIC/Mercyhurst standardization) is considered important in specific domains (e.g., intelligence analysis~\cite{heuer}) but other applications argue for direct access to the original numeric information.
    
    \item \textbf{Human performance aspects:} Understanding the characteristics of narratives that lead to improved human performance is an ongoing research problem\cite{metoyer2018coupling}.
    Narratives provide increased comprehension, interest, and engagement and are known to  contribute ``distinct cognitive pathways of comprehension'' with increased recall, ease of comprehension, and shorter reading times~\cite{Dahlstrom_2014}.
    Conversely, the challenge of the written word implies slowness and error-prone behavior due to short-term memory limits. 
\end{itemize}

In general, successful narrative research requires a standardization of both the generation and evaluation space, and an understanding of how a narrative fits into the larger comprehension process of the analyst.

%% file: 04-formative.tex
\section{Crowdsourced Study: Narration for Causality}
\label{sec:crowdsourced-study}

We conducted a crowdsourced study
to \change{change 12}\aj{understand how narratives augment causal data exploration through visual analysis.}

\subsection{Participants}

\change{change 13} \aj{We recruited our participants through crowdsourcing from \textit{Amazon Mechanical Turk (MTurk)} to complete visual analysis tasks that did not require prior training or data visualization expertise.
Owing to the nature of MTurk, we had limited control over participant demographics, technology, and skill level.
However, prior work indicates that simple tasks such as ours are flexible to a crowdsourced study design~\cite{Heer2010}.
We planned to recruit 150 participants; all were drawn from within the United States due to tax and compensation restrictions by our Institutional Review Board (IRB).
To ensure that our participants understood our task instructions, we screened our participants for working English knowledge.
Participants were allowed to participate only once.
We estimated our study completion time to be 20--30 minutes, and compensated our participants ethically at a rate of at least \$8/hour (similar to the U.S.\ federal minimum wage in 2019 of \$7.25).  
}

\subsection{Apparatus}

We required our participants to use a desktop computer (no mobile devices), and the study was distributed through a web browser.
We ensured that the visual representations, textual narratives, and their labels were legible for all common device formats.
The testing platform was implemented as a \textit{Qualtrics} survey with static trials saved as non-interactive \change{change 14}\aj{mockups that were manually created using \textit{Microsoft PowerPoint}, and were based on various factors such as polarity of links, link overlaps, and the number of intervening/objective nodes. The narratives were created manually based Section~\ref{sec:pipeline} (Figure~\ref{fig:task}).}

\subsection{Experimental Factors}
\label{sec:experimental-factors}

\change{change 15} \aj{Our goal was to first experimentally understand how the presence of a narrative augments causal analysis using visual representation.
We chose a more familiar and less temporal causal representation (Causal Graph) and less familiar and more temporal causal representation (Hasse Diagram).}
We modeled four factors in our experiment:

\begin{itemize}
    \setlength{\itemsep}{0pt}
    \item\textbf{Causality Visualization (VR)}:
    The visual representation used for conveying causality.
    We chose two levels: \vspace{-0.25cm}
    \begin{itemize}
        \setlength{\itemsep}{0pt}
        \item \textbf{Causal Graph (NL)}: A Causal Graph is a node-link representation of the causal network.
        \item \textbf{Hasse Diagram (HD)}: We use a similar representation of Hasse diagrams as seen in previous work \cite{elmqvist2004animated}.
    \end{itemize}  
    \item\textbf{Textual Narrative (TN)}: This is a key factor in our study: the 1) presence \textbf{(ON)} or 2) absence \textbf{(OFF)} of a textual narrative.

    \item\textbf{Difficulty (DL)}: The difficulty of the trial is expressed in the size of the causal system involved in the event sequence.
    We chose three levels for this factor: 
    \vspace{-0.25cm}
    \begin{itemize}
        \setlength{\itemsep}{0pt}
        \item \textbf{Simple (S)}: 3 or 5 nodes, and up to 4 time-hops (T1\textendash T4)
        \item \textbf{Medium (M)}: 5 to 8 nodes, and up 5 time-hops (T1\textendash T5)
        \item \textbf{Hard (H)}: 9 to 12 nodes, and up to 5 time-hops (T1\textendash T5)
        
        We settled on these values through pilot testing to ensure that our tasks can typically be completed in 30 minutes. 
    \end{itemize}
    
    \item\textbf{Narrative Scope (NS)}: Inherently, the Hasse diagram affords the explicit showing of changes in a node across time intervals (e.g. T1\textendash T2; T2\textendash T3, etc.).
    On the other hand, the Causal Graph (NL) requires the user to follow the causal path between nodes to extrapolate temporal information.
    This means that the accompanying textual narrative could describe effect propagation between successive time-hops---\textbf{Instantaneous} (IS)---or can provide a \textbf{Cumulative} (CU) summary across all observed time.

\end{itemize}

\begin{figure*}[tbh]
    \centering
    \includegraphics[width=\textwidth]{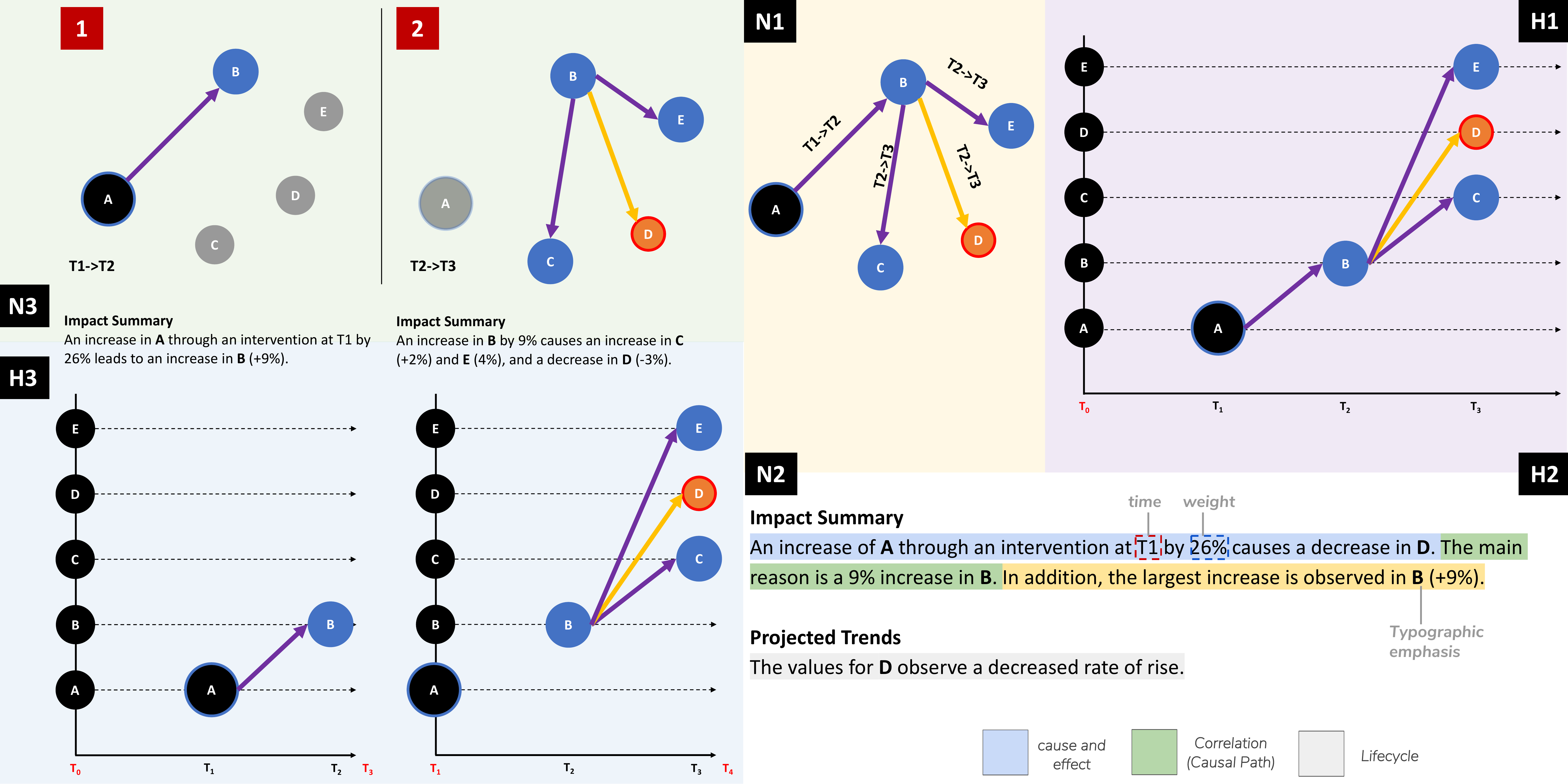}
    \caption{\change{change 38} \aj{Sample stimuli (DL: Simple) used in our 6 groups, with Hasse Diagram (HD) and Causal Graphs (NL). Sample narrative from groups N2 and H2 has been annotated with elements used from our design space.}}
    \label{fig:task}
\end{figure*}

This leads to 24 conditions.
Since NS is only relevant for situations when TN is ON, this yields a total of 18 conditions.
We presented a total of 12 causal graph systems (CGS) to each of our participants. \
\change{change 16} \aj{Although we do not include all aspects of our design space as experimental conditions, we use our narrative rendering pipeline in our mockups.
We were also limited by the non-interactivity of our stimulus.
Figure~\ref{fig:task} shows a representation of the above mentioned factors, and also is annotated with applicable aspects of our narrative rendering pipeline.}
\change{change 17 and 18} \aj{The nodes and edges in both visual representations of our abstract data were drawn manually with the aim of reducing edge crossing and length minimization.
For larger and realistic datasets, we recommend using graph layout algorithms that minimize edge length and crossings.
We can also note that the generated narratives are similar to those in Figure~\ref{fig:narrative-example}, which are a manifestation of the proposed design space.}

\subsection{Experimental Design}

We used a mixed design in our study: between-subjects for VR, TN and NS; and within-subjects for DL (Table~\ref{table:EX-design}).

\begin{table}[htb]
    \centering
    \caption{Six groups with 25 participants per group; N=150 (25 $\times$ 6).}
    \begin{tabular}{|l|cccccc|}
        \hline
        \rowcolor[HTML]{C0C0C0} 
        \cellcolor[HTML]{000000} & \textbf{H1} & \textbf{H2} & \textbf{H3} & \textbf{N1} & \textbf{N2} & \textbf{N3} \\ \hline
        \cellcolor[HTML]{EFEFEF}\textbf{VR} & HD & HD & HD & NL & NL & NL \\ \hline
        \cellcolor[HTML]{EFEFEF}\textbf{TN} & OFF & ON & ON & OFF & ON & ON \\ \hline
        \cellcolor[HTML]{EFEFEF}\textbf{NS} & -- & CU & IS & -- & CU & IS \\ \hline
        \cellcolor[HTML]{EFEFEF}\textbf{DL} & \begin{tabular}[c]{@{}c@{}}S\\ M\\ H\end{tabular} & \begin{tabular}[c]{@{}c@{}}S\\ M\\ H\end{tabular} & \begin{tabular}[c]{@{}c@{}}S\\ M\\ H\end{tabular} & \begin{tabular}[c]{@{}c@{}}S\\ M\\ H\end{tabular} & \begin{tabular}[c]{@{}c@{}}S\\ M\\ H\end{tabular} & \begin{tabular}[c]{@{}c@{}}S\\ M\\ H\end{tabular} \\ \hline
    \end{tabular}
    \label{table:EX-design}
\end{table}

Each participant saw all conditions of difficulty, but only one causality visualization.
The relatively small total number of conditions enabled us to keep the session duration shorter than 30 minutes in duration to minimize fatigue and maximize attention for crowdworkers.
In total, we planned to recruit 25 participants to each of six groups.

\subsection{Analysis Tasks}

Causal systems are complex structures that involve many  processes (events) and messages propagating through a network of connections.
In our user study, we use the words `node' to mean a process and `link' to mean a connection between `nodes'.
The comprehensibility of a cause-effect relationship between, say, two nodes might also require an understanding of other effects that have propagated or will propagate through the system. 
Broadly, an understanding of \textit{causality} might require a user to ask questions such as a) what factors caused a specific effect?, b) how does the effect on a node affect other connected nodes and to what extent?, c) what are the sequential and temporal impacts of this effect on the entire system?, d) how does this effect change or not change the earlier trend of the node.
We adapted types of analysis tasks from previous work \cite{elmqvist2004animated}, and created 24 tasks for our participants.

Each task was of one of the following three types \textbf{(QT)}:  1) Influence analysis \textbf{(I)}, 2) Cause-effect analysis \textbf{(C)}, and 3) Life-cycle Analysis \textbf{(L)}. 
\change{change 19} \aj{These task types are adaptations of causality information described in Section~\ref{sec:stepCausalityExtraction}}.
In the tutorial, we explained to our participants that we choose certain \textit{Intervention} and \textit{Objective} nodes to analyze causal relationships.
Table~\ref{table:tasks} shows the 9 types of \change{change 20}\aj{tasks included in our study.}

\change{change 21}\aj{Within each group, each participant saw 12 graph systems (4 $\times$ [S, M, H]).
A fixed order of increasing graph difficulty and tasks were used to improve familiarity by limiting chances of early task failure.
Each graph system had 2 analysis sub-tasks.
We distributed the first 8 task sub-types sequentially to each graph system within a DL, and alternated Trend (L4) with Spike (L5), in the event of a particular graph system showing a spike in a particular node (more on spikes in Section~\ref{ss:summarizing_causality}).
Thus, each DL covered all the Task types (QT).
Each sub-task required participants to read the question text and choose 1 out of 4 possible responses.
Thus, for 150 participants, we planned to collect a total of 3,600 trials \textendash  150 $\times$ 2 (questions) $\times$ 4 (graph systems) $\times$ 3 (DL).}

\aj{These graph systems were modeled after abstract causal relationships with each node being labeled by alphabets (Figure~\ref{fig:task}).
We avoided modelling real-world phenomena to avoid knowledge bias affecting performance.
As described in Section~\ref{sec:experimental-factors}, we assume that all 4 repetitions of a DL are equally simple or hard.
Each survey page consisted of a chart (VR + TN + NS) corresponding to the experimental group.}

\subsection{Collected Metrics}

The tasks for all trials were controlled so that all participants saw the same graph systems, and were asked the same set of questions to allow comparison of participant performance between experimental groups. 

\paragraph{Performance Measures.}

$Correctness$ (\textsc{true} or \textsc{false}) is our primary performance measure to interpret the effectiveness of narratives in augmenting visual exploration.
We also recorded time spent on each trial (from when the two tasks were displayed until the participant submitted both the answers) to understand if and how \textit{Completion Time} influences correctness.
However, due to a limitation in Qualtrics and the need to maintain low session time, we recorded both tasks together.

\paragraph{Subjective Responses.}
\aj{We also asked our participants to rate the ease-of-understanding of both graphs, and narratives (when applicable) after each DL.}
This was measured on a 5-point Likert scale (1: extremely easy, 5: extremely difficult).
In the conditions where TN was ON, we also asked participants to rate the usefulness of the narratives on a 5-point scale (1: extremely useful, 5: not at all useful).
Participants also provided open-ended feedback about graphs and narratives.

\begin{table*}[tbh]
    \centering
    \begin{tabular}{lll}
    \toprule
    \textbf{Task type (QT)} & \textbf{Task sub-type} & \textbf{Question Structure} \\
    \midrule
     & Major Cause (I1)& Considering all the nodes, which node(s) caused the most influence on the system? \\
    \multirow{-2}{*}{Influence (I)} & Most Affected (I2) & Considering all the nodes, which node(s) were affected the most by changes in the system? \\ \midrule
     & Cause-Effect (C1) & Which statement best describes the cause-effect relationship between \textless{}I\textgreater{} and \textless{}O\textgreater{}? \\ 
    \multirow{-2}{*}{Causality (C)} & Major Factors (C2) & Choose all the nodes, including the objective that were affected by a change in \textless{}I\textgreater{}. \\ \midrule
     & Max Increase (L1) & Excluding interventions/objectives, which node(s) goes through the greatest increase? \\
     & Max Decrease (L2) & Excluding interventions/objectives, which node(s) goes through the greatest decrease? \\
     & Time-Change (L3) & In the above system, at which time does node \textless{}X\textgreater{} increase/decrease the most from its initial level? \\ 
     & Trend (L4)& Which statement best describes the trend that node \textless{}X\textgreater{} goes through? \\
    \multirow{-5}{*}{Lifecycle (L)} & Spike (L5)& In the above system, which node goes through a sharp increase or decrease? \\ \bottomrule
    \end{tabular}
    \vspace{-0.25cm}
    \caption{List of task types and corresponding question structures for our user study.
    Each trial corresponded to a given task sub-type.}
    \label{table:tasks}
\end{table*}

\subsection{Procedure}

All recruitment was conducted via MTurk.
Participants that fit the eligibility criteria opened the Qualtrics survey in a separate browser window.
At the end of their participation, they copied a unique completion code back into the MTurk interface, and were later paid as their work was checked. 
Each session started with a consent form with waived signed consent.
Failing to give consent terminated the experiment.
Participants were instructed that they could abandon their session at any time.
Unfortunately, due to constraints in Qualtrics, we were not able to pay participants who only completed a partial session.

After consenting, we showed participants a tutorial to explain the visualization and narrative that they would see.
We also explained causal relationships, and how to interpret them.
The tutorial included 3 examples of simple causal relationships.
There was also a separate page explaining the visual mappings that were used in our visualizations.
This legend information was also accessible across every survey page, along with the visualization and a sample narrative from the tutorial.

Then participants were shown a single illustrated page of instructions explaining the task.
Additionally, we also introduced 3 ``attention trials,'' which involved 3 easy cause-effect (C) tasks.
Each DL had one attention trial.
The purpose of these attention trials was to eliminate responses from crowdworkers who did not pay attention to the task and only ``clicked through'' the experiment.
Participants that spent less than 7 minutes (roughly 1/3rd the average study duration) were also discarded; these participants were also not paid.
We informed participants in the consent form that they would be paid only after response validation.


Typical sessions lasted between 7 to 50 minutes in duration.
A few participants took significantly longer to complete their sessions, but our logs indicate that these participants took significant breaks between trials (presumably due to interruptions halfway through).
Some participants also contacted us with reasons for the delay, such as trials genuinely being hard, or network issues.
We believe that the effective time spent on the experiment was no more than 23 minutes.
Participants were also asked some demographic questions about their age, education level, and knowledge of statistical concepts and graph visualizations.



\begin{figure}[ht]
    \centering
    \includegraphics[width=\linewidth]{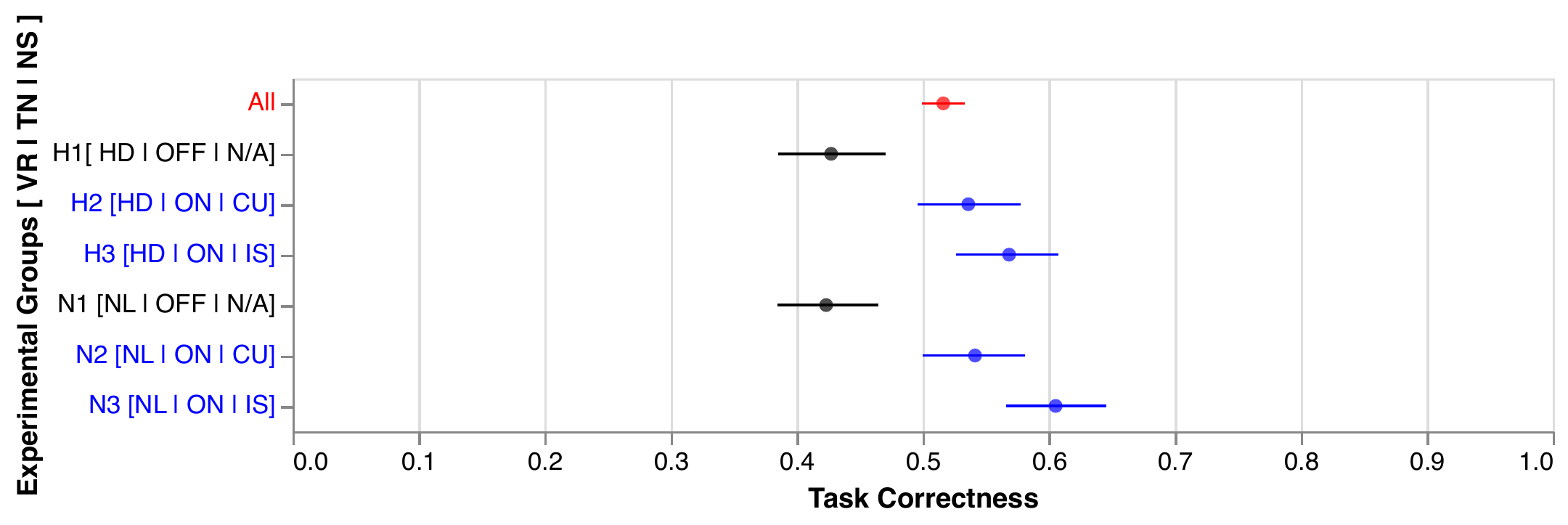}
    \caption{Correctness comparison between 6 experimental groups (error bars represent 95\% confidence intervals derived through bootstrapping).}
    \label{study_correctness}
\end{figure}
\vspace{-0.4cm}
\subsection{Results}

We ran our crowdsourced graphical perception study on MTurk and collected a total of 4,824 responses from 201 unique respondents.
This was higher than the 150 we had initially planned.
During the recruitment process, we invalidated and rejected respondents (n=44) that just ``clicked through'' and completed the survey in less than 7 minutes.
Thus, 157 participants were compensated for their time.
During our analysis, we excluded data from participants (n=20) who spent less than 10 minutes on the survey.
We expected a reasonable attempt to take 20 minutes based on our pilot, and believe that our complex perception tasks along with the tutorial required at least half of the estimated time.
We also eliminated one participant that submitted a survey response after 3.8 hours.
We present below results from the analysis of n=134 participants that completed 3,216 tasks (trials).
The trials were distributed across experimental groups as follows: H1-480 (n=20) | H2-528 (n=22) | H3-528 (n=22) | N1-600 (n=25) | N2-552 (n=23) | N3-528 (n=22).
\change{change 22} \aj{We analyzed all our data using estimation methods to derive 95\% confidence intervals (CIs).
We employed non-parametric bootstrapping~\cite{efron1992bootstrap} with R = 1,000 iterations.
This was done to follow current best practices for fair statistics in the field of HCI~\cite{Dragicevic2016}.}

\vspace{-0.1cm}
\paragraph{Task Correctness.}
Overall, we observed an accuracy of 51.6\% (1,661/3,216) (Figure~\ref{study_correctness}).
First, we observed that the participants assigned to experimental groups that included a textual narrative  | H2, H3, N2, N3 | outperformed those assigned to groups without a narrative | H1, N1.
Secondly, we noted that participants that interacted with Causal Graphs (N1, N2, N3) performed better than those that used Hasse Diagrams (H1, H2, H3).
We believe this to be a byproduct of participants being more familiar with node-link diagrams.
We can infer that narratives providing Instantaneous NS (H3, N3) fare better across both VR, with the causal graphs (N3) outperforming all other groups.

\begin{figure}[ht]
    \centering
    \includegraphics[width=\linewidth]{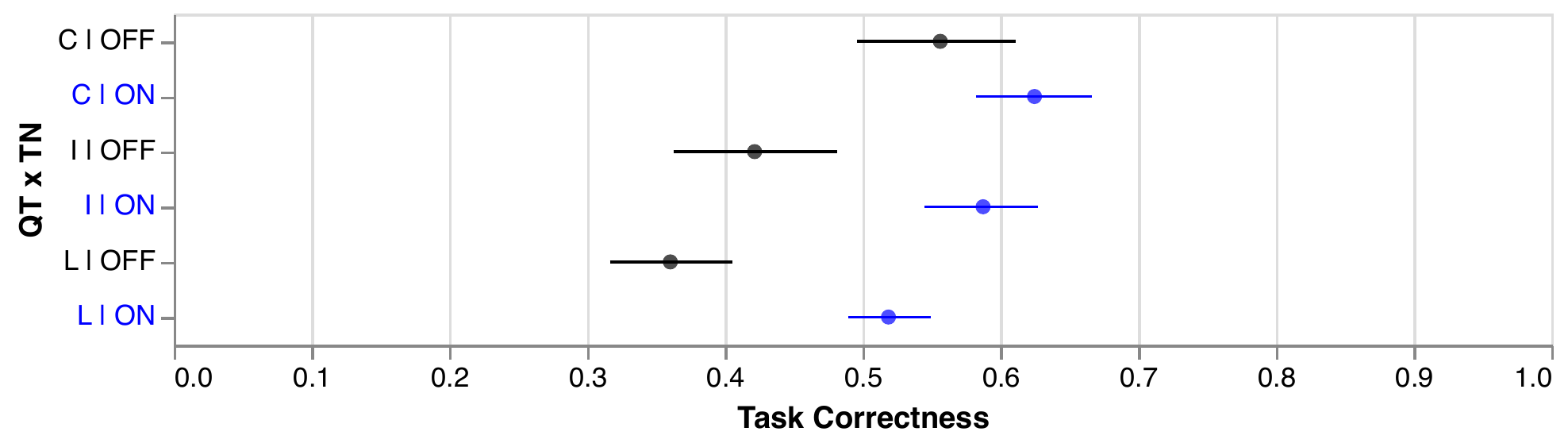}
    \caption{Comparing effectiveness of the narrative in answering different question types  (error bars represent 95\% confidence intervals).}
    \label{fig:correctnes_questiontype}
\end{figure}

Figure~\ref{fig:correctnes_questiontype} highlights the specific type of questions that the narratives were most effective in answering.
Although the correctness increases for each condition that includes narratives, participants found the presence of narratives most helpful in answering the Influence (\textbf{I}) and Life-cycle Analysis (\textbf{L}) questions, for both type of visualizations.
This became a useful insight while deciding on the modules in Section~\ref{sec:system}.

\begin{figure}[ht]
    \centering
    \includegraphics[width=\linewidth]{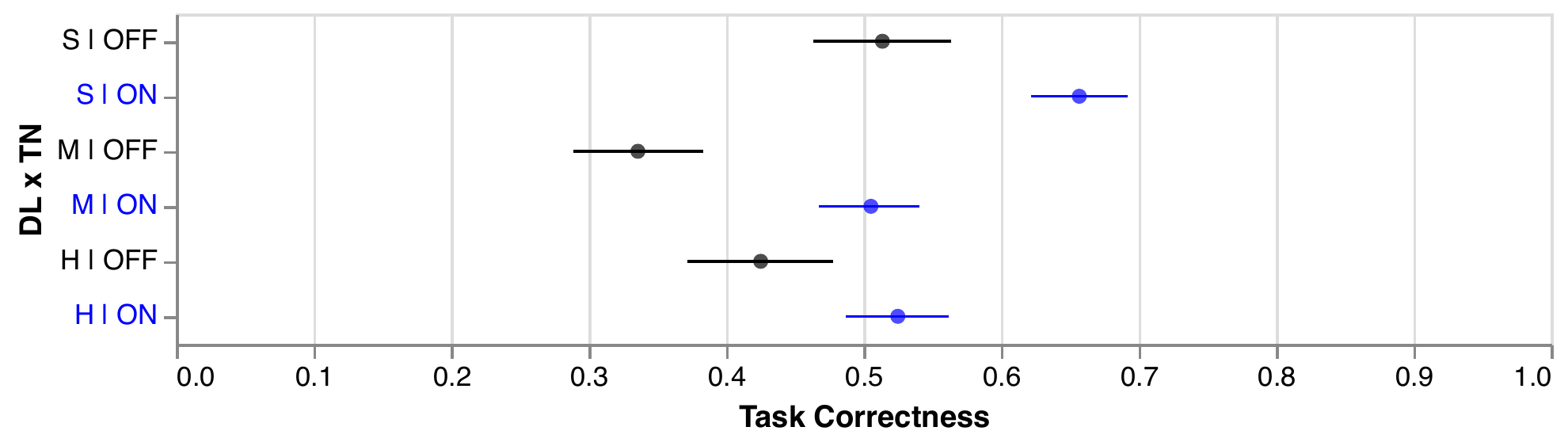}
    \caption{Comparing the correctness between graph difficulty levels (error bars represent 95\% confidence intervals).}
    \label{fig:correctness_difficulty}
\end{figure}

Figure~\ref{fig:correctness_difficulty} further highlights the improvement in the correctness of participant's scores, across each difficulty level (S, M, H), when the visualizations are coupled with textual narratives. \change{change 23} The comparatively lower task correctness improvements for the Hard (H) task type, in comparison to the Simple (S) and Medium (M) graph sets, can be attributed to the inherent added complexity, in terms of the added edges and number of nodes, within those datasets.

\begin{figure}[ht]
    \centering
    \includegraphics[width=\linewidth]{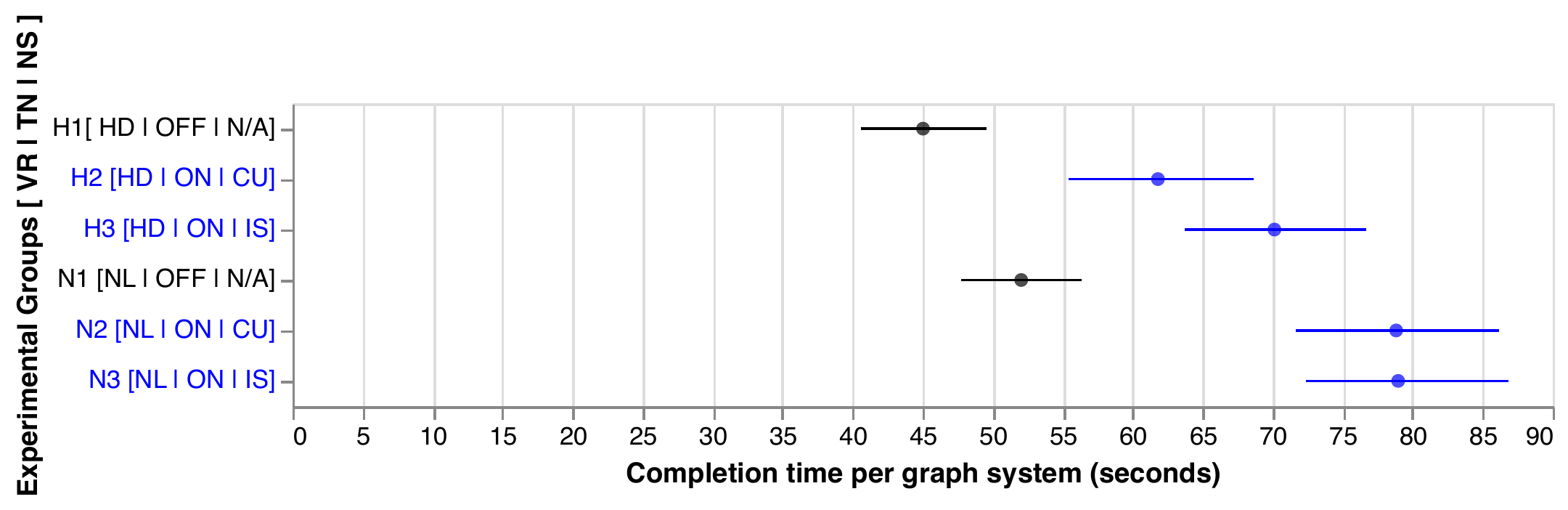}
    \caption{Completion time for the test trials across different conditions (error bars represent 95\% confidence intervals).}
    \label{fig:time}
\end{figure}

\vspace{-0.1cm}
\paragraph{Completion Time.}

Completion time was measured per graph system.
There were 4 repetitions of graph systems for every DL | S, M, H.
In other words, completion time reflects time spent by a participant for two tasks.
We eliminated outlier trials three standard deviations away from the mean for our analysis (Figure~\ref{fig:time}).
We note that participants took much longer in groups where a narrative was present (\aj{H2, H3}, N2, N3); with participants taking more time for Causal Graphs.
Participants who were provided with narratives (H2, H3, N2, N3) took on an average 23.6 seconds more to answer 2 analysis tasks per graph as compared to participants without the narrative. 

\paragraph{Subjective Responses.}

On an average, our participants ranked ease-of-understanding of the DL in the following order: Simple (mean=3.21), Medium (mean=3.68), Hard (mean=3.95).
Graphs were rated as more easily understandable in the conditions where textual narratives were present versus when narratives were absent:
Simple (mean=3.55 [OFF] vs. mean=3.0 [ON]); Medium (mean=3.82 [OFF] vs. mean=3.60 [ON]); Hard (mean=3.55 [OFF] vs. mean=3.0[ON]). 
The same trend was observed in the ease-of-understanding of narratives for each DL: Simple (mean=2.98), Medium (mean=3.38), Hard (mean=3.80).
Additionally, the usefulness of the narratives decreased with increasing graph difficulty:
Simple (mean=2.22), Medium (mean=3.38), Hard (mean=3.05).

Reviewing open-ended feedback showed us that difficulty understanding was attributed to unfamiliarity with a visualization: ``\textit{It was somewhat challenging because I'm not familiar with this type of graph.}
Additionally, the abstract nature of our graph systems, and also novelty effects influenced difficulty.
P111 says, ``\textit{Their abstract nature was the most difficult to understand.
Observing this with a real life example would make it easier to visualize and conceptualize.}''
Finally, our participants indicated that they used both visualization and narratives for causal inference: \change{change 25} \aj{``\textit{I think having both the summary and the color coded chart makes it much easier to understand [...].}''}


\subsection{Discussion}

The main takeaway from our crowdsourced study is that narratives complement visualizations by providing descriptions to explain changes in the causal system. \change{change 26} \aj{Based on our analysis of difficulty level and subjective responses, we believe that narratives will be more useful as the complexity of the system increases.
Our participants also indicated that interactivity would have eased task difficulty---a known limitation in our study.
We also strongly believe that interactivity can be leveraged to facilitate details on-demand in the narratives, especially when system complexity is bound to increase verbosity.} \aj{The fact that Causal Graphs had a higher accuracy score than Hasse Diagrams further corroborated the prioritization of DAGs during the DOI step, and drove us to use them as the visualization medium in the \textsc{CauseWorks} system.
The study also encourages us to allocate a separate paragraph to talk about the trends followed by important nodes, owing to the high accuracy gains observed in the Lifecycle (\textbf{L}) task.}

We also observed that the experimental groups that had higher accuracy also demonstrated higher completion times.
We believe that the additional time stems from having to read the narratives before making an inference. \change{change 27} \aj{This supports the aggregation DOI prioritization feature, wherein to reduce the verbosity of the textual snippets, nodes experiencing similar trends should be combined.}
This corroborates with our prediction that narratives aid in causal inference by providing descriptive texts that explain the changes occurring in the network.

%% file: 05-application.tex
\begin{figure*}[tbh]
    \includegraphics[width=\textwidth]{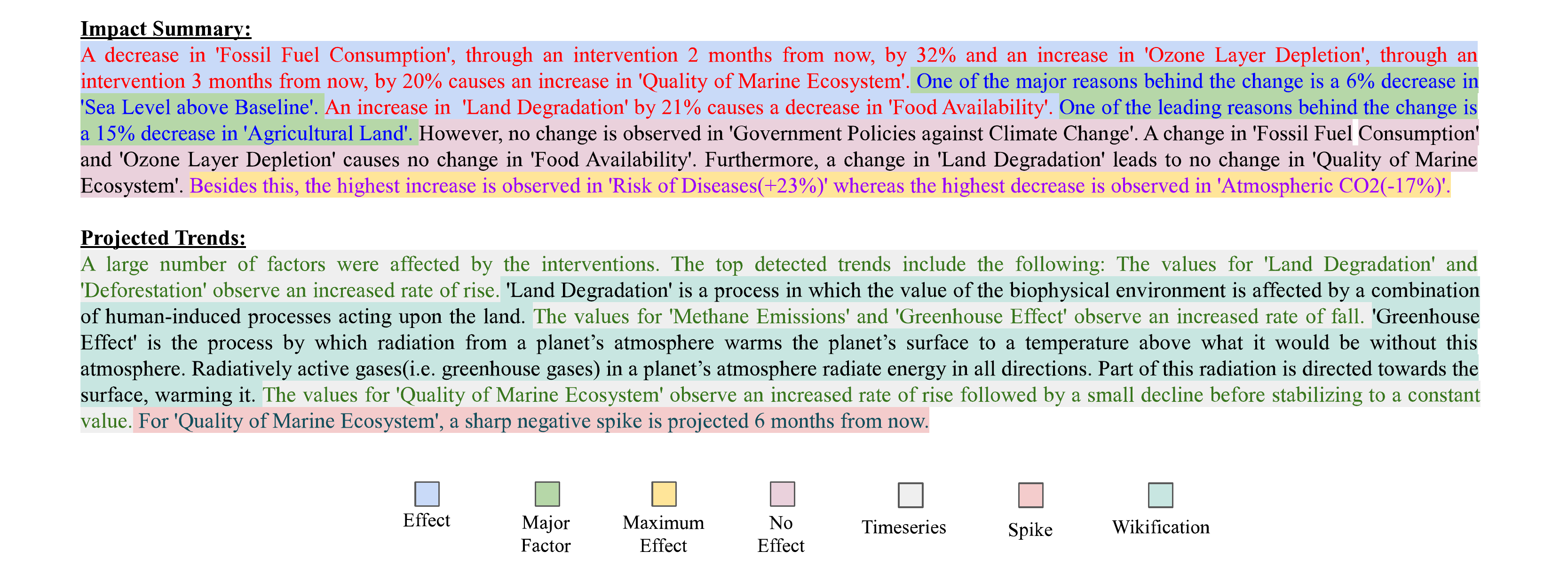}
    \vspace{-0.3cm}
    \caption{Narrative explaining a detailed causal network. \change{change 31} \aj{The `processes' or `events' are depicted within single-quotes in this figure}.}
    \label{comparison}
\end{figure*}

\section{Application: The CauseWorks System}
\label{sec:system}

\change{change 28} \aj{\textsc{CauseWorks}, a system for intelligence analysis~\cite{10.1145/1124772.1124890,8107978,doi:10.1080/10447318.2014.905422}, integrates a range of network analysis, natural language generation (NLG), and data analytics techniques to develop coherent, concise, and explainable causal visualizations augmented by narratives for use by analysts.}
Drawing on our design space, our visualizations and narratives provide two main mechanisms (Figure~\ref{comparison}): (1) a summary of changes and their impact on the objectives; and (2) additional projected trends.

\begin{figure*}[ht]%
 \centering
 \subfloat[Screenshot of the \textsc{CauseWorks} system.]{\includegraphics[width=\linewidth, height=7.5cm]{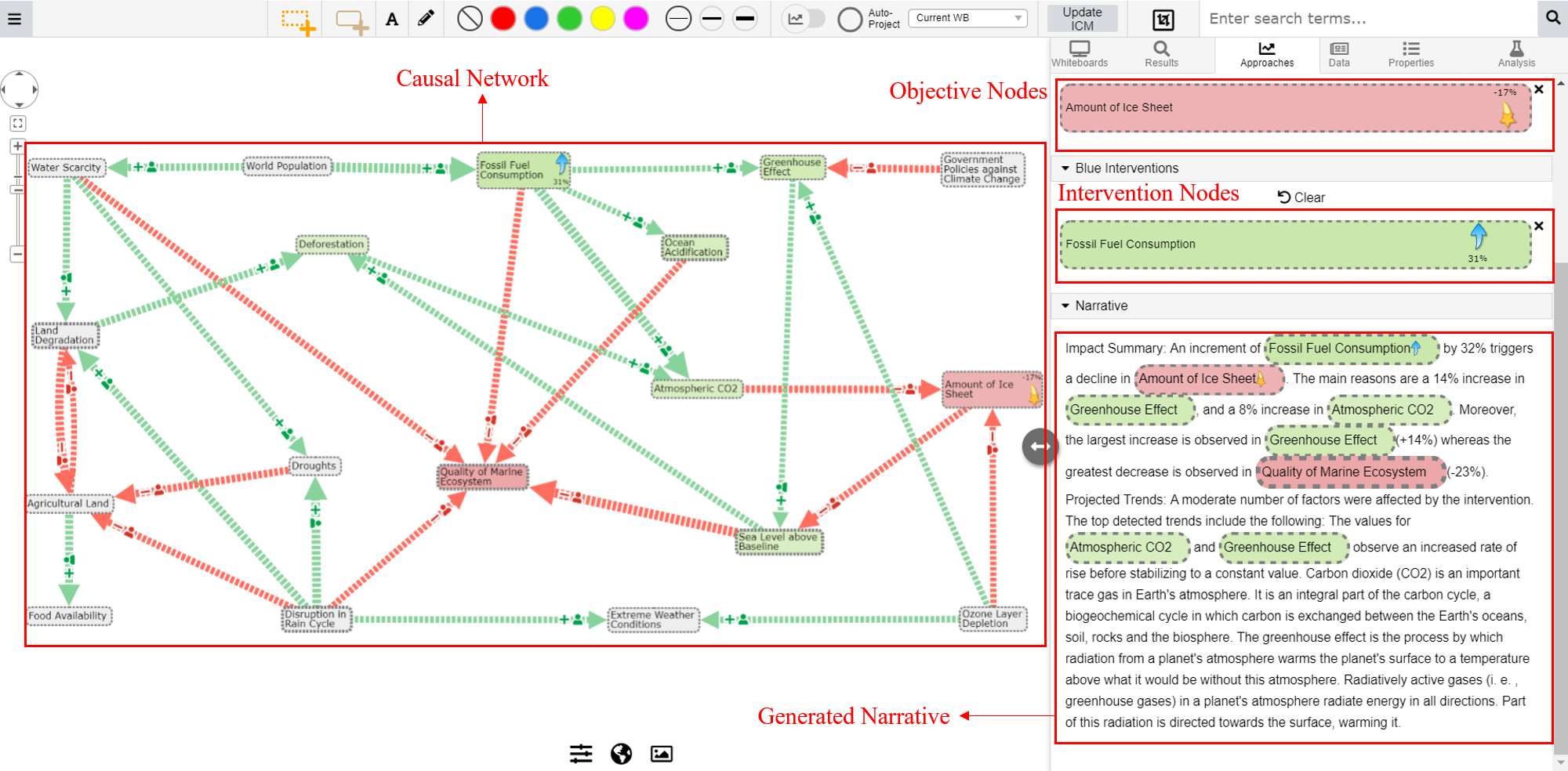}\label{toolOverview}}\\
 \subfloat[The use of color gradients to signify the polarity and impact of the effects with a darker shade corresponding to a higher impact.]{\includegraphics[ width=0.45\textwidth,height=2cm]{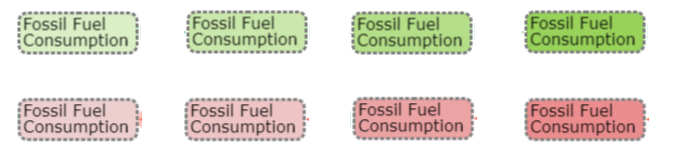}\label{nodeColor}}%
 \hspace{5mm}
 \subfloat[The use of edge thickness to signify the effect propagation capability of the edges, with a thicker line corresponding to a larger effect propagation.]{\includegraphics[height=2cm,width=0.45\textwidth]{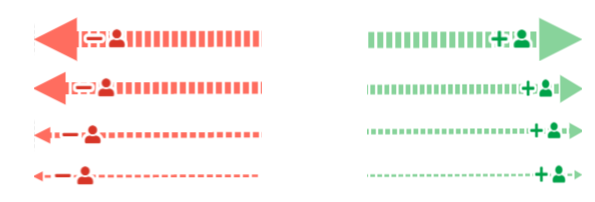}\label{linkThickness}}\\
 \caption{Overview of interface and features of the \textsc{CauseWorks} system.}%
\end{figure*}

\subsection{System Overview}

Figure~\ref{toolOverview} shows a screenshot of the \textsc{CauseWorks} system. \change{change 29} \aj{Various visual aspects of the design space (Section~ \ref{sec:renderingNarratives} and \ref{sec:interactingNarratives}), such as `Node Coloring', `Hyperlinking' and `Brushing', form an integral part of the narrative rendering process and its subsequent interactivity. 
Moreover, the performance metrics helped determine the usefulness of the various types of information snippets (Section~\ref{sec:stepCausalityExtraction}), whereas the subjective responses played an important role in deciding the order in which the various information snippets  are bundled together (Section~\ref{sec:stepCalculatingOrder}).}
The left pane displays the \textit{whiteboard}, a drawing space for causal graphs that allows the user to create and edit the network by adding, deleting, and modifying nodes as well as the edges amongst nodes, thus defining the semantics of the network.
The \textit{whiteboard} itself is unbounded, which allows the pane to incorporate a large number of nodes, and can be navigated using the \textit{scrolling wheel} and the \textit{magnifying scope} tools on the top left.
Furthermore, the system also displays the chosen \textit{objective} nodes, \textit{intervention} nodes as well as the generated \textit{narrative}, the placeholders for which can be seen in the right pane. \change{change 30}\aj{The interactive GUI allows the user to select multiple intervention nodes and multiple objective nodes, and displays an explanatory narrative in real-time.}

\subsection{Extracting Causality}
\label{ss:summarizing_causality}

The impact summary elucidates how interventions introduced over one or more nodes propagate through the network and change target nodes (\textit{Effect}) and the major nodes that help propagate that change (\textit{Major Effect}).
Note that the interventions could be made over one or multiple source nodes, and, further, they could be point interventions introduced at a specific timestamp or a sustained intervention introduced over a time period.
The precise differences in how such interventions create observable changes in the target nodes is dependent on the causal model semantics (e.g., whether it is an ODE-based model or a discrete time-stamped Bayesian model), which is beyond the scope of this paper.
Irrespective of the causal semantics, the impact summary encapsulates the cumulative effect of the interventions and identifies nodes in the causal path that depict the highest and least changes.

\paragraph{Impact Summary.}

Generating a summary of causal impacts is non-trivial due to the multitude of paths between source and target nodes.
An effective narrative must reduce the number of words utilized to describe the associated effects.
Below we detail how changes made on a set of interventions propagate through the network and affect the target nodes (\textit{Effect}), the major nodes that help propagate that change (\textit{Major Effect}), the intervening nodes that have no observable variation on the target nodes (\textit{No Effect}) and, finally, the nodes that experience the most impact (\textit{Max Effect}).

\definecolor{effect}{HTML}{c9daf8}
\definecolor{majorfactor}{HTML}{b6d7a8}
\definecolor{noeffect}{HTML}{ead1dc}
\definecolor{maxeffect}{HTML}{ffe599}
\definecolor{timeseries}{HTML}{efefef}
\definecolor{spike}{HTML}{f4cccc}
\definecolor{wikipedia}{HTML}{c8e6e1}

\vspace{-0.02in}

\begin{itemize}
\setlength{\itemsep}{0pt}
\item \colorbox{effect}{\textbf{Effect Module:}}
The `Effect Module' usually contributes the first sentence of the narrative and provides information on the propagation effect of each intervention on the specified target nodes.
The set of source nodes and target nodes are grouped together based on common nodes in their causal path via a dictionary of $\langle key,value \rangle$ pairs.
Then, paths are grouped by merging source nodes that have at least one common node for each target node and then subsequently merging together target nodes.
This often requires multiple passes and merge operations over the dictionary constructed.
Figure~\ref{comparison} depicts a sample snippet detailing the effect of decreasing \textit{Fossil Fuel Consumption} on \textit{Quality of Marine Ecosystem}.

\item \colorbox{majorfactor}{\textbf{Major Effect Module:}}
Each `Effect' sentence in the above module may or may not be followed by a sentence from the `Major Effect' module. This module tries to capture the important nodes along the causal path between the set of source and target nodes, thus shining light on those causal path nodes that experienced the highest variation in either direction. This enlightens the user regarding the nodes that were the highest contributors to the effect propagation. Sample text snippets are shown in Figure~\ref{comparison} to list out the major factors that propagate the effect between the chosen intervention and target nodes.

\item\colorbox{noeffect}{\textbf{No Effect Module:}}
This module articulates the specific source nodes that aren't responsible for the change observed in the target node as well as
the target nodes that remain unaffected due to the combined effect of all the interventions imposed on the network. This step can
potentially lead to highly verbose
sentences, thus affecting readability.
To address this problem we introduce
another grouping over the
(input, output) node pairs, create
n-grams of the source nodes, and 
articulate the
most frequently occurring tuples amongst all the target nodes. 
These groups of tuples are
then plugged into sentences, which are then included in the narrative. 
Figure~\ref{comparison} shows sample text snippets showing the non-impact of the interventions on `Government Policies against Climate Change' as well as describes the $\langle Intervention, Objective \rangle$ pairs that are not connected.

\item\colorbox{maxeffect}{\textbf{Maximum Effect Module:}}
The narrative generated until now focuses only on a subset of the whole network. This subset covers the edges and the nodes that lie in the causal paths between the set of \textit{Intervention} and \textit{Objective} nodes. However, there may still be nodes that might have been affected by the interventions but may have not been considered before. These nodes may provide interesting insights to the user and thus are worth adding to the final narrative. Hence, this module traverses through all the nodes in the system, instead of only the causal path nodes, and finds the nodes experiencing the maximum variation along both the positive and negative axis. Finally, it wraps both the nodes in a well structured sentence and attaches it to the end of the \textit{Impact Summary} narrative. Figure~\ref{comparison} points out the most positively impacted node, \textit{(Risk of Diseases)}, as well as the most negatively impacted node, \textit{(Atmospheric $CO_2$)}.

\end{itemize}

\vspace{-0.3cm}
\paragraph{Projected Trends.}

While the \textit{Impact Summary} articulates the overall influence of the source nodes on the target nodes, it leaves out information such as the temporal patterns observed by the nodes, or spikes in values that may have occurred in the course of the intervention, or other contextual information from external data sources (e.g., Wikipedia). We outline these parts of the narrative below.

\begin{itemize}
\setlength{\itemsep}{0pt}
\vspace{-0.25cm}
\item\colorbox{timeseries}{\textbf{Time Series Module:}}
The Time Series module parses over the temporal information for entities in the causal path between the source and target nodes and captures key change trajectories observed over those nodes. Following a k-means clustering~\cite{kmeans}
over the temporal progressions \change{change 32}\aj{across a 12 month period} (with number
of clusters selected using the
Silhouette coefficient~\cite{shilocoff}), the clusters
are sorted based on the number of nodes in
each cluster and the high volume clusters
are verbalized in the narrative. To
limit the description length, a Pagerank
score~\cite{pagerank} is used as a filtering criterion
to determine the most important nodes.
Figure~\ref{comparison} details the time series patterns observed by `Land Degradation', `Deforestation', `Methane Emissions' and `Greenhouse Effect'.

\item\colorbox{spike}{\textbf{Spike Detection Module:}}
Important nodes found in the previous step are further analyzed to check for presence of spikes or troughs during the timespan in consideration. This provides information to the user of any key abnormalities or milestones that might have occurred in these nodes. Each sentence of the \textit{Time Series module} may or may not be followed by an output 
from the `Spike Detection' module.
For detecting spikes in the timeseries, the concept of a moving window is used to
distinguish between gradual and sudden
rises or falls in the time series value.
Example text snippet showing the spikes in the time series value for `Quality of Marine Ecosystem' is shown in Figure~\ref{comparison}.

\item\colorbox{wikipedia}{\textbf{Wikification Module:}}
This constitutes the final module of the generated narrative. However, similar to the scenario with previous module, the \textit{Wikification} module is also interleaved with the \textit{Time Series} module to provide a seamless and continuous reading experience to the user.
This module involves parsing through the \change{change 33} \aj{summary paragraphs in the corresponding} Wikipedia pages for important nodes mentioned in the \textit{Timeseries} module and attaching key descriptive information to provide context
to the narrative.
If a Wikipedia page does not exist for the node mentioned, this module is skipped in its entirety.

\end{itemize}

\vspace{-0.35cm}
\subsection{Generating Textual Narratives}

Figure~\ref{comparison} shows an example narrative\footnote{Please refer to our supplementary material for the reference graph.}.
Most importantly, the narrative highlights important aspects of the cumulative causal impacts caused by the interventions on the specified objectives.
It also explains the effect that the interventions made on `Fossil Fuel Consumption(-31\%)', `Land Degradation(+21\%)` and `Ozone Layer Depletion(+20\%)` had on the target nodes, `Quality of Marine Ecosystem`, `Food Availability` and `Government Policies against Climate Change`.

Beyond the basic cause and effect relationships, the narrative also accounts for the changes to the entire system by mentioning the nodes experiencing the highest rise and decline across the whole network.
Furthermore, it clusters together nodes experiencing the same value patterns, and details the time-series patterns as well as spikes for the important nodes in those clusters.
It also provides additional insights, by attaching the associated Wikipedia summary, for the available nodes.

\vspace{-0.03in}
\subsection{Expert Review: \textsc{CauseWorks} Narratives}
\label{ss:expert_review}

We conducted an expert review~\cite{DBLP:journals/cga/ToryM05} to validate the narrative engine.

\paragraph{Method.}

We recruited 5 experts who worked with causal systems in varying capacities (P1 and P5 were developers working on building system frameworks and analytics for causal systems; P2 and P4 were usability experts working on user research and visualization design for causal systems; and P3 was a visualization expert working with causal systems).
We engaged with our experts in an hour-long semi-structured feedback session in a remote video call where they indirectly interacted with the \textsc{CauseWorks} system.
We encouraged the participants to think aloud, and interrupt at any point
to ask questions, make, and share observations.
These experts all work with planners who model and infer causality with the aid of visualization.
P1, P3, and P5 were familiar with causal visualization and modelling, but were relatively unfamiliar with narrative generation.
We gave a brief tutorial that explained both the visualization and the narrative structure.
We created a scenario to demonstrate the system features.
Two researchers collaborated with experts in the review: one researcher and the expert performed pair-analytics\cite{DBLP:conf/hicss/Arias-HernandezKGF11} while exploring the scenario, while the second researcher asked questions.
We explained to our experts that the focus of the session was to critique the generated narratives in the system, and not features such as the visualization or other user-interface elements. 

\vspace{-0.03in}

\paragraph{Results.} 

Overall, our experts were impressed with how the narrative augments the causal graph in the system, especially to tackle large-scale causal systems ``\textit{with multiple factors}'' (P2), and the narrative ``\textit{puts everything in context}'' (P2) when beginning causal exploration.
\change{change 34} \aj{Referring to our design space,} we summarize these results below:

\vspace{-0.2cm}

\begin{itemize}
    \setlength{\itemsep}{0pt}
    \item \textbf{Content Generation:}

    P3 suggested making the narrative more robust by including a `model summary' of the underlying causal model: ``\textit{Narrative should try (very) hard to have model scope (temporal and geographic scope)}.''
    Experts also noted that the narrative compensates for information loss from visual mapping by showing absolute values.
    Said P2: ``\textit{[my] first instinct was to look at the graph because that would tell the specific percentage}.''
    P1 noted that the reliability of the narrative can be improved by ``\textit{including the baseline trends}.''

    \item\textbf{Document Structuring and Aggregation:} Experts were satisfied with the presentation order of causal information.
    P3 commented that the order of presentation can remain the same for a particular type of narrative and may change if more types are introduced.
    For example, the impact summary, projected trends, and model summary may have different causal information---aggregated and structured in different order.

    \item\textbf{Realization and Interaction:} All 5 experts acknowledged the idea of a character budget, and that the current rendering of the textual narratives can be improved with the use of rendering effects such as hierarchical lists (all 5) and interactivity such as Brushing (P2, P4), Search (P5), and Sliders (P5).
\end{itemize}

%% file: 06-discussion.tex
\section{Discussion}
\label{sec:discussion}

\aj{Our work indicates that textual narratives help} users infer information from causal networks. 
More specifically, the user study shows our method faring favorably for time, correctness, and confidence of information absorption.
These narratives also reinforce the `narrative intelligence' viewpoint proposed by Blair and Meyer~\cite{blair1997tools}.
Narratives can be used to generate quick, precise, and informative reports (or subsections of reports) \change{change 35}\aj{~\cite{LATIF201927}} owing to their structured representations. 

The \textsc{CauseWorks} system adds another layer of abstraction to narratives.
Furthermore, the use cases presented here support past findings on visualization rhetoric~\cite{hullman2011visualization} in combining interactivity with organized information presentations to enhance the decision-making process for the end-user.
The findings are also in line with graph comics~\cite{bach2016telling}, which explored the effectiveness of using textual snippets with graphical images for communicating changes in dynamic networks.

\change{change 36} \aj{As we mentioned earlier, we did not experimentally validate all of our design space in the crowdsourced study.
However, we used our 4-step narrative rendering pipeline broadly in both the mockups of the crowdsourced study (e.g., causality information extraction, typographic emphasis, calculating order, and font size) and \textsc{CauseWorks} (causality information extraction, calculating order, textual rendering and color, word-scale graphics, and interactivity through brushing).
Our expert review results also validate our design choices.}

\aj{We believe that future evaluations on the effectiveness of our narrative design space can help expand the space for causal systems, and eventually other systems.
To test this hypothesis, we also plan on conducting another user study with the \textsc{CauseWorks} system to evaluate the performance benefits offered by the system in a more interactive setting.}
Limitations in our work pertain to the scope of questions about causality that it can answer, e.g., we have not focused on communicating dynamics of the system as a whole. \change{change 37} \aj{The current methodology is also designed in-line with the temporal datasets. Future work will be targeted towards a more generic approach that ingests non-temporal datasets as well as incorporating a visual DOI function to focus the user's attention on important nodes and links.}

%% file: 07-conclusion.tex
\vspace{-0.2cm}
\section{Conclusion} 
\label{sec:conclusion}

We have presented a review of the design space for a specialized form of data-driven storytelling: the use of natural language narratives for causal network data.
Based on the review, we isolated several interesting questions about the role of textual narratives for this purpose.
To answer these questions, we conducted a large-scale crowdsourced user study where participants saw causal systems of increasing complexity.
The data was displayed using one of two visualization techniques, causal graphs and Hasse diagrams, with and without the presence of textual narratives.
The main finding is that the coupling of causality visualization techniques with textual narratives significantly increases accuracy and acts as a pivotal complement to information visualization.

%% file: appendix.tex
\clearpage
\appendix
\newpage
\pagenumbering{arabic}

\onecolumn
\setcounter{page}{1}
\setcounter{section}{0}
\setcounter{figure}{0}
\renewcommand\thefigure{\Alph{figure}}

\begin{center}
\huge{\textsf{\thetitle}}

\vspace{1cm}

\Large\textsf{Supplementary Materials}
\end{center}

\section{Additional Figures}

\begin{figure*}[tbh]
    \centering
    \includegraphics[width=\textwidth]{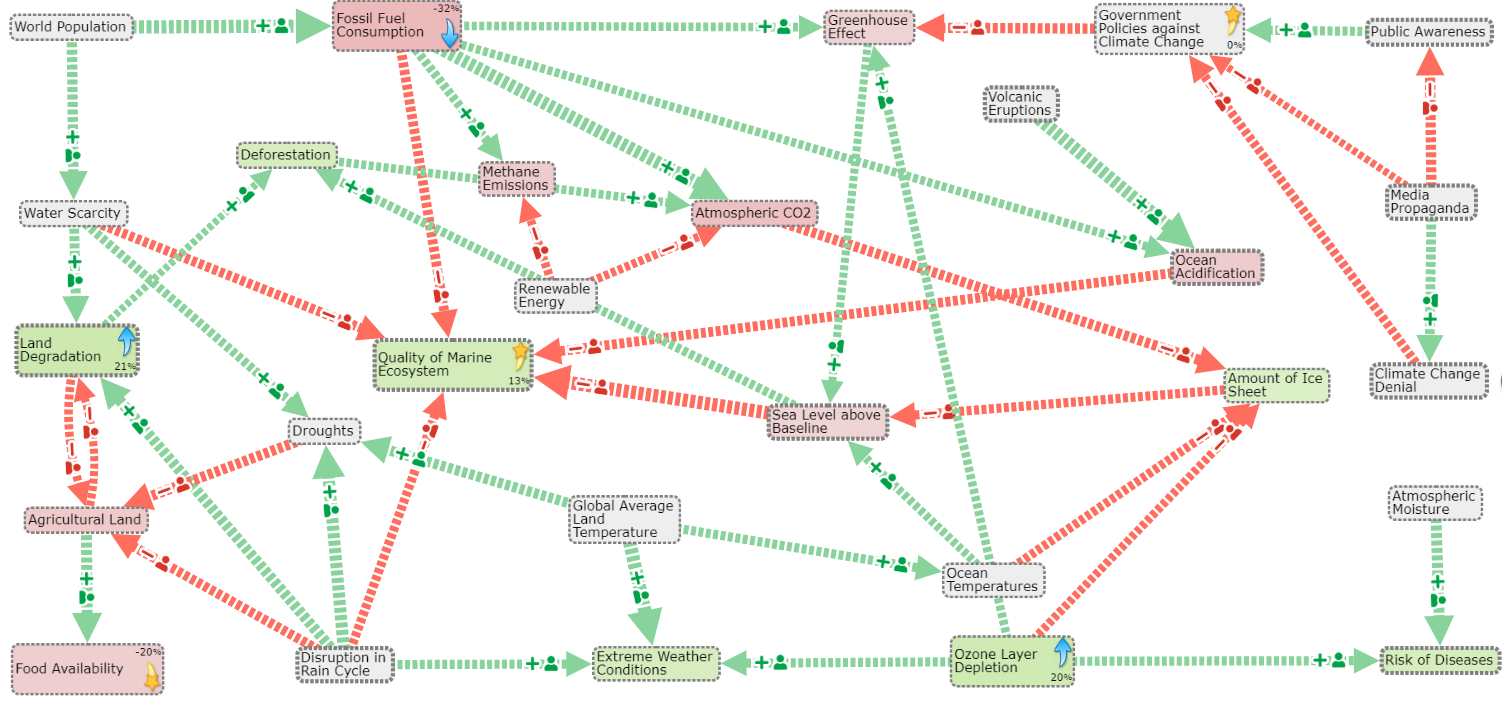}
    \caption{Sample causal network visualized in \textsc{CauseWorks}.}
    \label{causalnetwork}
\end{figure*}

\section{Video}

We have attached a companion video showcasing the \textsc{CauseWorks} system.
\section{Graph Samples}

We have also attached images of the graph systems that we used in our crowdsourced study. 

The naming convention is: \textbf{[Experimental condition][Difficulty Level][Repetition]}

Example: [H1][N][1] | The file name will be `H1N1.png'

\section{Anonymized Performance Data}

We have attached anonymized correctness, time, and subjective responses.

%% file: casual-narrative.bbl
\begin{thebibliography}{10}

\bibitem{causeex}
{DARPA}, {C}ausal {E}xploration,
  \url{https://www.darpa.mil/program/causal-exploration}, 2018.

\bibitem{DBLP:conf/hicss/Arias-HernandezKGF11}
R.~Arias{-}Hern{\'{a}}ndez, L.~T. Kaastra, T.~M. Green, and B.~D. Fisher.
\newblock Pair analytics: Capturing reasoning processes in collaborative visual
  analytics.
\newblock In {\em Proceedings of the Hawaii International International
  Conference on Systems Science}, pp. 1--10. {IEEE} Computer Society, 2011.
  doi: {{%
10\hspace{.1pt}\discretionary{.}{%
}{.}\hspace{.4pt}1109\discretionary{/}{%
}{/}HICSS\hspace{.1pt}\discretionary{.}{%
}{.}\hspace{.4pt}2011\hspace{.1pt}\discretionary{.}{%
}{.}\hspace{.4pt}339}}


\bibitem{bach2016telling}
B.~Bach, N.~Kerracher, K.~W. Hall, S.~Carpendale, J.~Kennedy, and
  N.~Henry~Riche.
\newblock Telling stories about dynamic networks with graph comics.
\newblock In {\em Proceedings of the ACM Conference on Human Factors in
  Computing Systems}, pp. 3670--3682, 2016. doi: {{%
10\hspace{.1pt}\discretionary{.}{%
}{.}\hspace{.4pt}1145\discretionary{/}{%
}{/}2858036\hspace{.1pt}\discretionary{.}{%
}{.}\hspace{.4pt}2858387}}


\bibitem{bae2017understanding}
J.~Bae, T.~Helldin, and M.~Riveiro.
\newblock Understanding indirect causal relationships in node-link graphs.
\newblock {\em Computer Graphics Forum}, 36(3):411--421, 2017. doi: {{%
10\hspace{.1pt}\discretionary{.}{%
}{.}\hspace{.4pt}1111\discretionary{/}{%
}{/}cgf\hspace{.1pt}\discretionary{.}{%
}{.}\hspace{.4pt}13198}}


\bibitem{Beck2014}
F.~Beck, M.~Burch, S.~Diehl, and D.~Weiskopf.
\newblock The state of the art in visualizing dynamic graphs.
\newblock In R.~Borgo, R.~Maciejewski, and I.~Viola, eds., {\em State of the
  Art Reports for the Eurographics Conference on Visualization}. Eurographics
  Association, 2014. doi: {{%
10\hspace{.1pt}\discretionary{.}{%
}{.}\hspace{.4pt}2312\discretionary{/}{%
}{/}eurovisstar\hspace{.1pt}\discretionary{.}{%
}{.}\hspace{.4pt}20141174}}


\bibitem{7864462}
F.~{Beck} and D.~{Weiskopf}.
\newblock Word-sized graphics for scientific texts.
\newblock {\em IEEE Transactions on Visualization and Computer Graphics},
  23(6):1576--1587, 2017. doi: {{%
10\hspace{.1pt}\discretionary{.}{%
}{.}\hspace{.4pt}1109\discretionary{/}{%
}{/}TVCG\hspace{.1pt}\discretionary{.}{%
}{.}\hspace{.4pt}2017\hspace{.1pt}\discretionary{.}{%
}{.}\hspace{.4pt}2674958}}


\bibitem{blair1997tools}
D.~Blair and T.~Meyer.
\newblock Tools for an interactive virtual cinema.
\newblock In {\em Creating Personalities for Synthetic Actors}, pp. 83--91.
  Springer, 1997. doi: {{%
10\hspace{.1pt}\discretionary{.}{%
}{.}\hspace{.4pt}1007\discretionary{/}{%
}{/}BFb0030572}}


\bibitem{Bunge1979CausalityAM}
M.~Bunge.
\newblock {\em Causality and Modern Science}.
\newblock Dover Publications, 1979.

\bibitem{chang2009defining}
R.~Chang, C.~Ziemkiewicz, T.~M. Green, and W.~Ribarsky.
\newblock Defining insight for visual analytics.
\newblock {\em IEEE Computer Graphics and Applications}, 29(2):14--17, 2009.
  doi: {{%
10\hspace{.1pt}\discretionary{.}{%
}{.}\hspace{.4pt}1109\discretionary{/}{%
}{/}MCG\hspace{.1pt}\discretionary{.}{%
}{.}\hspace{.4pt}2009\hspace{.1pt}\discretionary{.}{%
}{.}\hspace{.4pt}22}}


\bibitem{10.1111:cgf.14034}
A.~Chatzimparmpas, R.~M. Martins, I.~Jusufi, K.~Kucher, F.~Rossi, and
  A.~Kerren.
\newblock {The State of the Art in Enhancing Trust in Machine Learning Models
  with the Use of Visualizations}.
\newblock {\em Computer Graphics Forum}, 2020. doi: {{%
10\hspace{.1pt}\discretionary{.}{%
}{.}\hspace{.4pt}1111\discretionary{/}{%
}{/}cgf\hspace{.1pt}\discretionary{.}{%
}{.}\hspace{.4pt}14034}}


\bibitem{Dahlstrom_2014}
M.~F. Dahlstrom.
\newblock Using narratives and storytelling to communicate science with
  nonexpert audiences.
\newblock {\em Proceedings of the National Academy of Sciences}, 111(Supplement
  4):13614--13620, 2014. doi: {{%
10\hspace{.1pt}\discretionary{.}{%
}{.}\hspace{.4pt}1073\discretionary{/}{%
}{/}pnas\hspace{.1pt}\discretionary{.}{%
}{.}\hspace{.4pt}1320645111}}


\bibitem{Diakopoulos2011}
N.~Diakopoulos, J.~DiMicco, J.~Hullman, K.~Karahalios, and A.~Perer.
\newblock Telling stories with data: The next chapter---a visweek 2011
  workshop, 2011.

\bibitem{DiMicco2010}
J.~DiMicco, M.~McKeon, and K.~Karahalios.
\newblock Telling stories with data---a {VisWeek} 2010 workshop, 2010.

\bibitem{Dragicevic2016}
P.~Dragicevic.
\newblock {\em Fair Statistical Communication in HCI}, pp. 291--330.
\newblock Springer International Publishing, Cham, 2016. doi: {{%
10\hspace{.1pt}\discretionary{.}{%
}{.}\hspace{.4pt}1007\discretionary{/}{%
}{/}978\discretionary{%
}{-}{-}3\discretionary{%
}{-}{-}319\discretionary{%
}{-}{-}26633\discretionary{%
}{-}{-}6\_13}}


\bibitem{efron1992bootstrap}
B.~Efron.
\newblock Bootstrap methods: another look at the jackknife.
\newblock In {\em Breakthroughs in statistics}, pp. 569--593. Springer, 1992.
  doi: {{%
10\hspace{.1pt}\discretionary{.}{%
}{.}\hspace{.4pt}1007\discretionary{/}{%
}{/}978\discretionary{%
}{-}{-}1\discretionary{%
}{-}{-}4612\discretionary{%
}{-}{-}4380\discretionary{%
}{-}{-}9\_41}}


\bibitem{Eisner2008}
W.~Eisner.
\newblock {\em Graphic Storytelling and Visual Narrative}.
\newblock W. W. Norton \& Company, New York, NY, USA, 2008.

\bibitem{Elmqvist2003b}
N.~Elmqvist and P.~Tsigas.
\newblock Causality visualization using animated growing polygons.
\newblock In {\em Proceedings of the IEEE Symposium on Information
  Visualization}, pp. 189--196, 2003. doi: {{%
10\hspace{.1pt}\discretionary{.}{%
}{.}\hspace{.4pt}1109\discretionary{/}{%
}{/}INFVIS\hspace{.1pt}\discretionary{.}{%
}{.}\hspace{.4pt}2003\hspace{.1pt}\discretionary{.}{%
}{.}\hspace{.4pt}1249025}}


\bibitem{Elmqvist2003a}
N.~Elmqvist and P.~Tsigas.
\newblock Growing squares: Animated visualization of causal relations.
\newblock In {\em Proceedings of the ACM Symposium on Software Visualization},
  pp. 17--26, 2003. doi: {{%
10\hspace{.1pt}\discretionary{.}{%
}{.}\hspace{.4pt}1145\discretionary{/}{%
}{/}774833\hspace{.1pt}\discretionary{.}{%
}{.}\hspace{.4pt}774836}}


\bibitem{elmqvist2004animated}
N.~Elmqvist and P.~Tsigas.
\newblock Animated visualization of causal relations through growing 2d
  geometry.
\newblock {\em Information Visualization}, 3:154--172, 07 2004. doi: {{%
10\hspace{.1pt}\discretionary{.}{%
}{.}\hspace{.4pt}1057\discretionary{/}{%
}{/}palgrave\hspace{.1pt}\discretionary{.}{%
}{.}\hspace{.4pt}ivs\hspace{.1pt}\discretionary{.}{%
}{.}\hspace{.4pt}9500074}}


\bibitem{Elwert201313GC}
F.~Elwert.
\newblock {\em Graphical Causal Models}, pp. 245--273.
\newblock 03 2013. doi: {{%
10\hspace{.1pt}\discretionary{.}{%
}{.}\hspace{.4pt}1007\discretionary{/}{%
}{/}978\discretionary{%
}{-}{-}94\discretionary{%
}{-}{-}007\discretionary{%
}{-}{-}6094\discretionary{%
}{-}{-}3\_13}}


\bibitem{Feng2018}
M.~Feng, C.~Deng, E.~M. Peck, and L.~Harrison.
\newblock The effects of adding search functionality to interactive
  visualizations on the web.
\newblock In R.~L. Mandryk, M.~Hancock, M.~Perry, and A.~L. Cox, eds., {\em
  Proceedings of the {ACM} Conference on Human Factors in Computing Systems},
  pp. 137:1--137:13. {ACM}, New York, NY, USA, 2018. doi: {{%
10\hspace{.1pt}\discretionary{.}{%
}{.}\hspace{.4pt}1145\discretionary{/}{%
}{/}3173574\hspace{.1pt}\discretionary{.}{%
}{.}\hspace{.4pt}3173711}}


\bibitem{Gambhir2016RecentAT}
M.~Gambhir and V.~Gupta.
\newblock Recent automatic text summarization techniques: a survey.
\newblock {\em Artificial Intelligence Review}, 47:1--66, 2016. doi: {{%
10\hspace{.1pt}\discretionary{.}{%
}{.}\hspace{.4pt}1007\discretionary{/}{%
}{/}s10462\discretionary{%
}{-}{-}016\discretionary{%
}{-}{-}9475\discretionary{%
}{-}{-}9}}


\bibitem{geiger1990d}
D.~Geiger, T.~Verma, and J.~Pearl.
\newblock d-separation: From theorems to algorithms.
\newblock In {\em Machine Intelligence and Pattern Recognition}, vol.~10, pp.
  139--148. Elsevier, 1990. doi: {{%
10\hspace{.1pt}\discretionary{.}{%
}{.}\hspace{.4pt}1016\discretionary{/}{%
}{/}B978\discretionary{%
}{-}{-}0\discretionary{%
}{-}{-}444\discretionary{%
}{-}{-}88738\discretionary{%
}{-}{-}2\hspace{.1pt}\discretionary{.}{%
}{.}\hspace{.4pt}50018\discretionary{%
}{-}{-}X}}


\bibitem{Gershon2001}
N.~D. Gershon and W.~Page.
\newblock What storytelling can do for information visualization.
\newblock {\em Communications of the ACM}, 44(8):31--37, 2001. doi: {{%
10\hspace{.1pt}\discretionary{.}{%
}{.}\hspace{.4pt}1145\discretionary{/}{%
}{/}381641\hspace{.1pt}\discretionary{.}{%
}{.}\hspace{.4pt}381653}}


\bibitem{Goffin2017}
P.~Goffin, J.~Boy, W.~Willett, and P.~Isenberg.
\newblock An exploratory study of word-scale graphics in data-rich text
  documents.
\newblock {\em {{IEEE} Transactions on Visualization and Computer Graphics}},
  23(10):2275--2287, 2017. doi: {{%
10\hspace{.1pt}\discretionary{.}{%
}{.}\hspace{.4pt}1109\discretionary{/}{%
}{/}TVCG\hspace{.1pt}\discretionary{.}{%
}{.}\hspace{.4pt}2016\hspace{.1pt}\discretionary{.}{%
}{.}\hspace{.4pt}2618797}}


\bibitem{Gottschall2012}
J.~Gottschall.
\newblock {\em The Storytelling Animal: How Stories Make Us Human}.
\newblock Mariner Books, New York, NY, USA, 2012. doi: {{%
10\hspace{.1pt}\discretionary{.}{%
}{.}\hspace{.4pt}1075\discretionary{/}{%
}{/}ssol\hspace{.1pt}\discretionary{.}{%
}{.}\hspace{.4pt}2\hspace{.1pt}\discretionary{.}{%
}{.}\hspace{.4pt}2\hspace{.1pt}\discretionary{.}{%
}{.}\hspace{.4pt}07bor}}


\bibitem{dep-networks}
D.~Heckerman, D.~M. Chickering, C.~Meek, R.~Rounthwaite, and C.~M. Kadie.
\newblock Dependency networks for inference, collaborative filtering, and data
  visualization.
\newblock {\em Journal of Machine Learning Resesearch}, 1:49--75, 2000. doi:
  {{%
10\hspace{.1pt}\discretionary{.}{%
}{.}\hspace{.4pt}1162\discretionary{/}{%
}{/}153244301753344614}}


\bibitem{Heer2010}
J.~Heer and M.~Bostock.
\newblock Crowdsourcing graphical perception: Using mechanical turk to assess
  visualization design.
\newblock In {\em Proceedings of the ACM Conference on Human Factors in
  Computing Systems}, pp. 203--212. ACM, New York, NY, USA, 2010. doi: {{%
10\hspace{.1pt}\discretionary{.}{%
}{.}\hspace{.4pt}1145\discretionary{/}{%
}{/}1753326\hspace{.1pt}\discretionary{.}{%
}{.}\hspace{.4pt}1753357}}


\bibitem{heuer}
R.~J. Heuer~Jr.
\newblock Analysis of competing hypotheses.
\newblock {\em Psychology of Intelligence Analysis}, pp. 95--110, 1999.

\bibitem{8933695}
F.~{Hohman}, A.~{Srinivasan}, and S.~M. {Drucker}.
\newblock {TeleGam}: Combining visualization and verbalization for
  interpretable machine learning.
\newblock In {\em Proceedings of the IEEE Conference on Visualization}, pp.
  151--155, 2019. doi: {{%
10\hspace{.1pt}\discretionary{.}{%
}{.}\hspace{.4pt}1109\discretionary{/}{%
}{/}VISUAL\hspace{.1pt}\discretionary{.}{%
}{.}\hspace{.4pt}2019\hspace{.1pt}\discretionary{.}{%
}{.}\hspace{.4pt}8933695}}


\bibitem{Hullman2011}
J.~Hullman and N.~Diakopoulos.
\newblock Visualization rhetoric: Framing effects in narrative visualization.
\newblock {\em IEEE Transactions on Visualization and Computer Graphics},
  17(12):2231--2240, 2011. doi: {{%
10\hspace{.1pt}\discretionary{.}{%
}{.}\hspace{.4pt}1109\discretionary{/}{%
}{/}TVCG\hspace{.1pt}\discretionary{.}{%
}{.}\hspace{.4pt}2011\hspace{.1pt}\discretionary{.}{%
}{.}\hspace{.4pt}255}}


\bibitem{hullman2011visualization}
J.~Hullman and N.~Diakopoulos.
\newblock Visualization rhetoric: Framing effects in narrative visualization.
\newblock {\em {{IEEE} Transactions on Visualization and Computer Graphics}},
  17(12):2231--2240, 2011. doi: {{%
10\hspace{.1pt}\discretionary{.}{%
}{.}\hspace{.4pt}1109\discretionary{/}{%
}{/}TVCG\hspace{.1pt}\discretionary{.}{%
}{.}\hspace{.4pt}2011\hspace{.1pt}\discretionary{.}{%
}{.}\hspace{.4pt}255}}


\bibitem{hullman2013deeper}
J.~Hullman, S.~Drucker, N.~H. Riche, B.~Lee, D.~Fisher, and E.~Adar.
\newblock A deeper understanding of sequence in narrative visualization.
\newblock {\em {{IEEE} Transactions on Visualization and Computer Graphics}},
  19(12):2406--2415, 2013. doi: {{%
10\hspace{.1pt}\discretionary{.}{%
}{.}\hspace{.4pt}1109\discretionary{/}{%
}{/}TVCG\hspace{.1pt}\discretionary{.}{%
}{.}\hspace{.4pt}2013\hspace{.1pt}\discretionary{.}{%
}{.}\hspace{.4pt}119}}


\bibitem{kadaba2007visualizing}
N.~Kadaba, P.~Irani, and J.~Leboe.
\newblock Visualizing causal semantics using animations.
\newblock {\em {{IEEE} Transactions on Visualization and Computer Graphics}},
  13:1254--61, 11 2007. doi: {{%
10\hspace{.1pt}\discretionary{.}{%
}{.}\hspace{.4pt}1109\discretionary{/}{%
}{/}TVCG\hspace{.1pt}\discretionary{.}{%
}{.}\hspace{.4pt}2007\hspace{.1pt}\discretionary{.}{%
}{.}\hspace{.4pt}70528}}


\bibitem{kent}
S.~Kent.
\newblock {\em Words of Estimative Probability}.
\newblock 1964.

\bibitem{koller-friedman}
D.~Koller and N.~Friedman.
\newblock {\em Probabilistic Graphical Models: Principles and Techniques -
  Adaptive Computation and Machine Learning}.
\newblock The MIT Press, 2009. doi: {{%
10\hspace{.1pt}\discretionary{.}{%
}{.}\hspace{.4pt}5555\discretionary{/}{%
}{/}1795555}}


\bibitem{Kosara2013}
R.~Kosara.
\newblock Story points in {T}ableau {S}oftware.
\newblock Keynote at Tableau Customer Conference, Sept. 2013.

\bibitem{LATIF201927}
S.~Latif and F.~Beck.
\newblock Interactive map reports summarizing bivariate geographic data.
\newblock {\em Visual Informatics}, 3(1):27 -- 37, 2019.
\newblock Proceedings of PacificVAST. doi: {{%
10\hspace{.1pt}\discretionary{.}{%
}{.}\hspace{.4pt}1016\discretionary{/}{%
}{/}j\hspace{.1pt}\discretionary{.}{%
}{.}\hspace{.4pt}visinf\hspace{.1pt}\discretionary{.}{%
}{.}\hspace{.4pt}2019\hspace{.1pt}\discretionary{.}{%
}{.}\hspace{.4pt}03\hspace{.1pt}\discretionary{.}{%
}{.}\hspace{.4pt}004}}


\bibitem{8440852}
S.~{Latif} and F.~{Beck}.
\newblock Vis author profiles: Interactive descriptions of publication records
  combining text and visualization.
\newblock {\em IEEE Transactions on Visualization and Computer Graphics},
  25(1):152--161, 2019. doi: {{%
10\hspace{.1pt}\discretionary{.}{%
}{.}\hspace{.4pt}1109\discretionary{/}{%
}{/}TVCG\hspace{.1pt}\discretionary{.}{%
}{.}\hspace{.4pt}2018\hspace{.1pt}\discretionary{.}{%
}{.}\hspace{.4pt}2865022}}


\bibitem{Leitch1986}
T.~M. Leitch.
\newblock {\em What Stories Are: Narrative Theory and Interpretation}.
\newblock Pennsylvania State University Press, University Park, PA, 1986.

\bibitem{kmeans}
J.~MacQueen et~al.
\newblock Some methods for classification and analysis of multivariate
  observations.
\newblock In {\em Proceedings of the Berkeley Symposium on Mathematical
  Statistics and Probability}, vol.~1, pp. 281--297. Oakland, CA, USA, 1967.

\bibitem{metoyer2018coupling}
R.~Metoyer, Q.~Zhi, B.~Janczuk, and W.~Scheirer.
\newblock Coupling story to visualization: Using textual analysis as a bridge
  between data and interpretation.
\newblock In {\em Proceedings of the ACM Conference on Intelligent User
  Interfaces}, pp. 503--507, 2018. doi: {{%
10\hspace{.1pt}\discretionary{.}{%
}{.}\hspace{.4pt}1145\discretionary{/}{%
}{/}3172944\hspace{.1pt}\discretionary{.}{%
}{.}\hspace{.4pt}3173007}}


\bibitem{moezzi2017using}
M.~Moezzi, K.~B. Janda, and S.~Rotmann.
\newblock Using stories, narratives, and storytelling in energy and climate
  change research.
\newblock {\em Energy Research \& Social Science}, 31:1--10, 2017. doi: {{%
10\hspace{.1pt}\discretionary{.}{%
}{.}\hspace{.4pt}1016\discretionary{/}{%
}{/}j\hspace{.1pt}\discretionary{.}{%
}{.}\hspace{.4pt}erss\hspace{.1pt}\discretionary{.}{%
}{.}\hspace{.4pt}2017\hspace{.1pt}\discretionary{.}{%
}{.}\hspace{.4pt}06\hspace{.1pt}\discretionary{.}{%
}{.}\hspace{.4pt}034}}


\bibitem{Nenkova2012ASO}
A.~Nenkova and K.~McKeown.
\newblock A survey of text summarization techniques.
\newblock In {\em Mining Text Data}, 2012. doi: {{%
10\hspace{.1pt}\discretionary{.}{%
}{.}\hspace{.4pt}1007\discretionary{/}{%
}{/}978\discretionary{%
}{-}{-}1\discretionary{%
}{-}{-}4614\discretionary{%
}{-}{-}3223\discretionary{%
}{-}{-}4\_3}}


\bibitem{north2006toward}
C.~North.
\newblock Toward measuring visualization insight.
\newblock {\em IEEE Computer Graphics \& Applications}, 26(3):6--9, 2006. doi:
  {{%
10\hspace{.1pt}\discretionary{.}{%
}{.}\hspace{.4pt}1109\discretionary{/}{%
}{/}MCG\hspace{.1pt}\discretionary{.}{%
}{.}\hspace{.4pt}2006\hspace{.1pt}\discretionary{.}{%
}{.}\hspace{.4pt}70}}


\bibitem{pagerank}
L.~Page, S.~Brin, R.~Motwani, and T.~Winograd.
\newblock The {PageRank} citation ranking: Bringing order to the web.
\newblock In {\em Proceedings of the ACM Conference on the World Wide Web},
  1999. doi: {{%
10\hspace{.1pt}\discretionary{.}{%
}{.}\hspace{.4pt}1\hspace{.1pt}\discretionary{.}{%
}{.}\hspace{.4pt}1\hspace{.1pt}\discretionary{.}{%
}{.}\hspace{.4pt}38\hspace{.1pt}\discretionary{.}{%
}{.}\hspace{.4pt}5427}}


\bibitem{Pearl2000CausalityMR}
J.~Pearl.
\newblock {\em Causality: Models, Reasoning and Inference}.
\newblock Cambridge University Press, USA, 2nd ed., 2009. doi: {{%
10\hspace{.1pt}\discretionary{.}{%
}{.}\hspace{.4pt}5555\discretionary{/}{%
}{/}1642718}}


\bibitem{nlg2}
E.~Reiter and R.~Dale.
\newblock Building applied natural language generation systems.
\newblock {\em Natural Language Engineering}, 3:57--87, 1997. doi: {{%
10\hspace{.1pt}\discretionary{.}{%
}{.}\hspace{.4pt}1017\discretionary{/}{%
}{/}S1351324997001502}}


\bibitem{Dale2000}
E.~Reiter and R.~Dale.
\newblock {\em Building Natural Language Generation Systems}.
\newblock Studies in Natural Language Processing. Cambridge University Press,
  2000. doi: {{%
10\hspace{.1pt}\discretionary{.}{%
}{.}\hspace{.4pt}1017\discretionary{/}{%
}{/}CBO9780511519857}}


\bibitem{Henry2018}
N.~H. Riche, C.~Hurter, N.~Diakopoulos, and S.~Carpendale.
\newblock {\em Data-Driven Storytelling}.
\newblock A. K. Peters, Ltd., USA, 1st ed., 2018.

\bibitem{doi:10.1080/10447318.2014.905422}
C.~Rooney, S.~Attfield, B.~L.~W. Wong, and S.~Choudhury.
\newblock Invisque as a tool for intelligence analysis: The construction of
  explanatory narratives.
\newblock {\em International Journal of Human–Computer Interaction},
  30(9):703--717, 2014. doi: {{%
10\hspace{.1pt}\discretionary{.}{%
}{.}\hspace{.4pt}1080\discretionary{/}{%
}{/}10447318\hspace{.1pt}\discretionary{.}{%
}{.}\hspace{.4pt}2014\hspace{.1pt}\discretionary{.}{%
}{.}\hspace{.4pt}905422}}


\bibitem{shilocoff}
P.~J. Rousseeuw.
\newblock Silhouettes: A graphical aid to the interpretation and validation of
  cluster analysis.
\newblock {\em Journal of Computational and Applied Mathematics}, 20:53 -- 65,
  1987. doi: {{%
10\hspace{.1pt}\discretionary{.}{%
}{.}\hspace{.4pt}1016\discretionary{/}{%
}{/}0377\discretionary{%
}{-}{-}0427\discretionary{%
}{(}{(}87\discretionary{)}{%
}{)}90125\discretionary{%
}{-}{-}7}}


\bibitem{Schank1995}
R.~C. Schank and R.~P. Abelson.
\newblock Knowledge and memory: The real story.
\newblock In J.~Robert S.~Wyer, ed., {\em Advances in Social Cognition},
  vol.~8, pp. 1--85. Lawrence Erlbaum Associates, Hillsdale, NJ, USA, 1995.

\bibitem{Segel2010}
E.~Segel and J.~Heer.
\newblock Narrative visualization: Telling stories with data.
\newblock {\em IEEE Transactions on Visualization and Computer Graphics},
  16(6):1139--1148, 2010. doi: {{%
10\hspace{.1pt}\discretionary{.}{%
}{.}\hspace{.4pt}1109\discretionary{/}{%
}{/}TVCG\hspace{.1pt}\discretionary{.}{%
}{.}\hspace{.4pt}2010\hspace{.1pt}\discretionary{.}{%
}{.}\hspace{.4pt}179}}


\bibitem{Sevastjanova2018Going-45045}
R.~Sevastjanova, F.~Beck, B.~Ell, C.~Turkay, R.~Henkin, M.~Butt, D.~A. Keim,
  and M.~El-Assady.
\newblock Going beyond visualization: Verbalization as complementary medium to
  explain machine learning models.
\newblock In {\em Workshop on Visualization for AI Explainability at IEEE VIS},
  2018.

\bibitem{shachter2013bayes}
R.~D. Shachter.
\newblock Bayes-ball: The rational pastime (for determining irrelevance and
  requisite information in belief networks and influence diagrams).
\newblock {\em arXiv preprint arXiv:1301.7412}, 2013. doi: {{%
10\hspace{.1pt}\discretionary{.}{%
}{.}\hspace{.4pt}5555\discretionary{/}{%
}{/}2074094\hspace{.1pt}\discretionary{.}{%
}{.}\hspace{.4pt}2074151}}


\bibitem{Sless1981}
D.~Sless.
\newblock {\em Learning and Visual Communication}.
\newblock Wiley, New York, NY, USA, 1981.

\bibitem{endert-vast}
A.~Srinivasan, S.~M. Drucker, A.~Endert, and J.~Stasko.
\newblock Augmenting visualizations with interactive data facts to facilitate
  interpretation and communication.
\newblock {\em {{IEEE} Transactions on Visualization and Computer Graphics}},
  25(1):672--681, 2018. doi: {{%
10\hspace{.1pt}\discretionary{.}{%
}{.}\hspace{.4pt}1109\discretionary{/}{%
}{/}TVCG\hspace{.1pt}\discretionary{.}{%
}{.}\hspace{.4pt}2018\hspace{.1pt}\discretionary{.}{%
}{.}\hspace{.4pt}2865145}}


\bibitem{fromendert}
A.~Srinivasan, H.~Park, A.~Endert, and R.~C. Basole.
\newblock Graphiti: Interactive specification of attribute-based edges for
  network modeling and visualization.
\newblock {\em {{IEEE} Transactions on Visualization and Computer Graphics}},
  24(1):226--235, 2017. doi: {{%
10\hspace{.1pt}\discretionary{.}{%
}{.}\hspace{.4pt}1109\discretionary{/}{%
}{/}TVCG\hspace{.1pt}\discretionary{.}{%
}{.}\hspace{.4pt}2017\hspace{.1pt}\discretionary{.}{%
}{.}\hspace{.4pt}2744843}}


\bibitem{Thomas2005}
J.~J. Thomas and K.~A. Cook.
\newblock {\em Illuminating the Path: The Research and Development Agenda for
  Visual Analytics}.
\newblock IEEE Computer Society, addrIEEECS, 2005.

\bibitem{DBLP:journals/cga/ToryM05}
M.~Tory and T.~M{\"{o}}ller.
\newblock Evaluating visualizations: Do expert reviews work?
\newblock {\em {IEEE} Computer Graphics and Applications}, 25(5):8--11, 2005.
  doi: {{%
10\hspace{.1pt}\discretionary{.}{%
}{.}\hspace{.4pt}1109\discretionary{/}{%
}{/}MCG\hspace{.1pt}\discretionary{.}{%
}{.}\hspace{.4pt}2005\hspace{.1pt}\discretionary{.}{%
}{.}\hspace{.4pt}102}}


\bibitem{Vansina1985}
J.~Vansina.
\newblock {\em Oral Tradition as History}.
\newblock University of Wisconsin Press, Madison, WI, USA, 1985. doi: {{%
10\hspace{.1pt}\discretionary{.}{%
}{.}\hspace{.4pt}2307\discretionary{/}{%
}{/}3601125}}


\bibitem{Viegas2006}
F.~Vi\'{e}gas and M.~Wattenberg.
\newblock Communication-minded visualization: A call to action.
\newblock {\em IBM Systems Journal}, 45(4):801--812, 2006. doi: {{%
10\hspace{.1pt}\discretionary{.}{%
}{.}\hspace{.4pt}1147\discretionary{/}{%
}{/}sj\hspace{.1pt}\discretionary{.}{%
}{.}\hspace{.4pt}454\hspace{.1pt}\discretionary{.}{%
}{.}\hspace{.4pt}0801}}


\bibitem{7192729}
J.~{Wang} and K.~{Mueller}.
\newblock The visual causality analyst: An interactive interface for causal
  reasoning.
\newblock {\em IEEE Transactions on Visualization and Computer Graphics},
  22(1):230--239, 2016. doi: {{%
10\hspace{.1pt}\discretionary{.}{%
}{.}\hspace{.4pt}1109\discretionary{/}{%
}{/}TVCG\hspace{.1pt}\discretionary{.}{%
}{.}\hspace{.4pt}2015\hspace{.1pt}\discretionary{.}{%
}{.}\hspace{.4pt}2467931}}


\bibitem{ware2013}
C.~Ware.
\newblock Perceiving complex causation through interaction.
\newblock In {\em Proceedings of the Symposium on Computational Aesthetics},
  pp. 29–--35. Association for Computing Machinery, New York, NY, USA, 2013.
  doi: {{%
10\hspace{.1pt}\discretionary{.}{%
}{.}\hspace{.4pt}1145\discretionary{/}{%
}{/}2487276\hspace{.1pt}\discretionary{.}{%
}{.}\hspace{.4pt}2487279}}


\bibitem{10.1145/1124772.1124890}
W.~Wright, D.~Schroh, P.~Proulx, A.~Skaburskis, and B.~Cort.
\newblock The sandbox for analysis: Concepts and methods.
\newblock In {\em Proceedings of the ACM Conference on Human Factors in
  Computing Systems}, pp. 801–--810. Association for Computing Machinery, New
  York, NY, USA, 2006. doi: {{%
10\hspace{.1pt}\discretionary{.}{%
}{.}\hspace{.4pt}1145\discretionary{/}{%
}{/}1124772\hspace{.1pt}\discretionary{.}{%
}{.}\hspace{.4pt}1124890}}


\bibitem{8107978}
W.~{Wright}, D.~{Sheffield}, and S.~{Santosa}.
\newblock Argument mapper: Countering cognitive biases in analysis with
  critical (visual) thinking.
\newblock In {\em Proceedings of the International Conference on Information
  Visualisation}, pp. 250--255, 2017.

\bibitem{xiong2019illusion}
C.~Xiong, J.~Shapiro, J.~Hullman, and S.~Franconeri.
\newblock Illusion of causality in visualized data.
\newblock {\em IEEE Transactions on Visualization and Computer Graphics},
  26(1):853--862, 2019. doi: {{%
10\hspace{.1pt}\discretionary{.}{%
}{.}\hspace{.4pt}1109\discretionary{/}{%
}{/}TVCG\hspace{.1pt}\discretionary{.}{%
}{.}\hspace{.4pt}2019\hspace{.1pt}\discretionary{.}{%
}{.}\hspace{.4pt}2934399}}


\end{thebibliography}
